\documentclass[journal,twoside,web]{ieeecolor}

\PassOptionsToPackage{table,xcdraw,HTML}{xcolor}

\usepackage{xcolor} 
\usepackage{tmi}
\usepackage{cite}
\usepackage{multirow}
\usepackage{booktabs}
\usepackage{array}
\usepackage{caption}
\usepackage{cite}
\usepackage{arydshln}

\usepackage{amsmath,amssymb,amsfonts}
\usepackage{algorithmic}
\usepackage{colortbl}
\usepackage{graphicx}
\usepackage{makecell}
\usepackage{textcomp}

\definecolor{revisioncolor}{RGB}{51,112,189}

\newcommand{\revision}[1]{\textcolor{revisioncolor}{#1}}  
\renewcommand{\revision}[1]{#1}

\def\BibTeX{{\rm B\kern-.05em{\sc i\kern-.025em b}\kern-.08em
    T\kern-.1667em\lower.7ex\hbox{E}\kern-.125emX}}
\begin{document}
\title{PGVMS: A Prompt-Guided Unified Framework for Virtual Multiplex IHC Staining with Pathological Semantic Learning}
\author{Fuqiang Chen,  Ranran Zhang, Wanming Hu, Deboch Eyob Abera, Yue Peng, Boyun Zheng, Yiwen Sun, Jing Cai and Wenjian Qin 
\thanks{
This work was supported in part by the Ministry of Science and Technology’s key research and development program (2023YFF0723400),  National Natural Science Foundation of China (No. 62271475), Shenzhen-Hong Kong Joint Lab on Intelligence Computational Analysis for Tumor lmaging (E3G111), Guangdong Youth Talent program (2024TQ08A386), and the Youth Innovation Promotion Association CAS (2022365). \textit{(Fuqiang Chen and Ranran Zhang contributed equally to this work.) (Corresponding author: Wenjian Qin.)}}
\thanks{Fuqiang Chen, Deboch Eyob Abera and Yue Peng are with the Shenzhen Institutes of Advanced Technology, Chinese Academy of Sciences and University of Chinese Academy of Sciences, Beijing, China.}
\thanks{Wanming Hu is with the Department of Pathology, Sun Yat-sen University Cancer Center, State Key Laboratory of Oncology in South China, Guangzhou, China.}%
\thanks{Jing Cai and Yiwen Sun are with the Hong Kong Polytechnic University, Hong Kong SAR, China (e-mail: jing.cai@polyu.edu.hk).}
\thanks{Boyun Zheng is with the Department of Electronic Engineering, The Chinese University of Hong Kong, Hong Kong SAR 999077, China}
\thanks{Wenjian Qin and Ranran Zhang are with the Shenzhen Institutes of Advanced Technology, Chinese Academy of Sciences, Shenzhen, China (e-mail: wj.qin@siat.ac.cn).}}

\maketitle
\begin{abstract}
Immunohistochemical (IHC) staining enables precise molecular profiling of protein expression, with over 200 clinically available antibody-based tests in modern pathology. However, comprehensive IHC analysis is frequently limited by insufficient tissue quantities in small biopsies. 
Therefore, virtual multiplex staining emerges as an innovative solution to digitally transform H\&E images into multiple IHC representations, yet current methods still face three critical challenges: (1) inadequate semantic guidance for multi-staining, (2) inconsistent distribution of immunochemistry staining, and (3) spatial misalignment across different stain modalities. To overcome these limitations, we present a prompt-guided framework for virtual multiplex IHC staining using only uniplex training data (PGVMS). Our framework introduces three key innovations corresponding to each challenge: First, an adaptive prompt guidance mechanism employing a pathological visual language model dynamically adjusts staining prompts to resolve semantic guidance limitations (Challenge 1). Second, our protein-aware learning strategy (PALS) maintains precise protein expression patterns by direct quantification and constraint of protein distributions (Challenge 2). Third, the prototype-consistent learning strategy (PCLS) establishes cross-image semantic interaction to correct spatial misalignments (Challenge 3).
Evaluated on two benchmark datasets, PGVMS demonstrates superior performance in pathological consistency. In general, PGVMS represents a paradigm shift from dedicated single-task models toward unified virtual staining systems.

\end{abstract}


\section{Introduction}
\label{sec:introduction}

\begin{figure}[t]
\centerline{\includegraphics[width=\columnwidth]{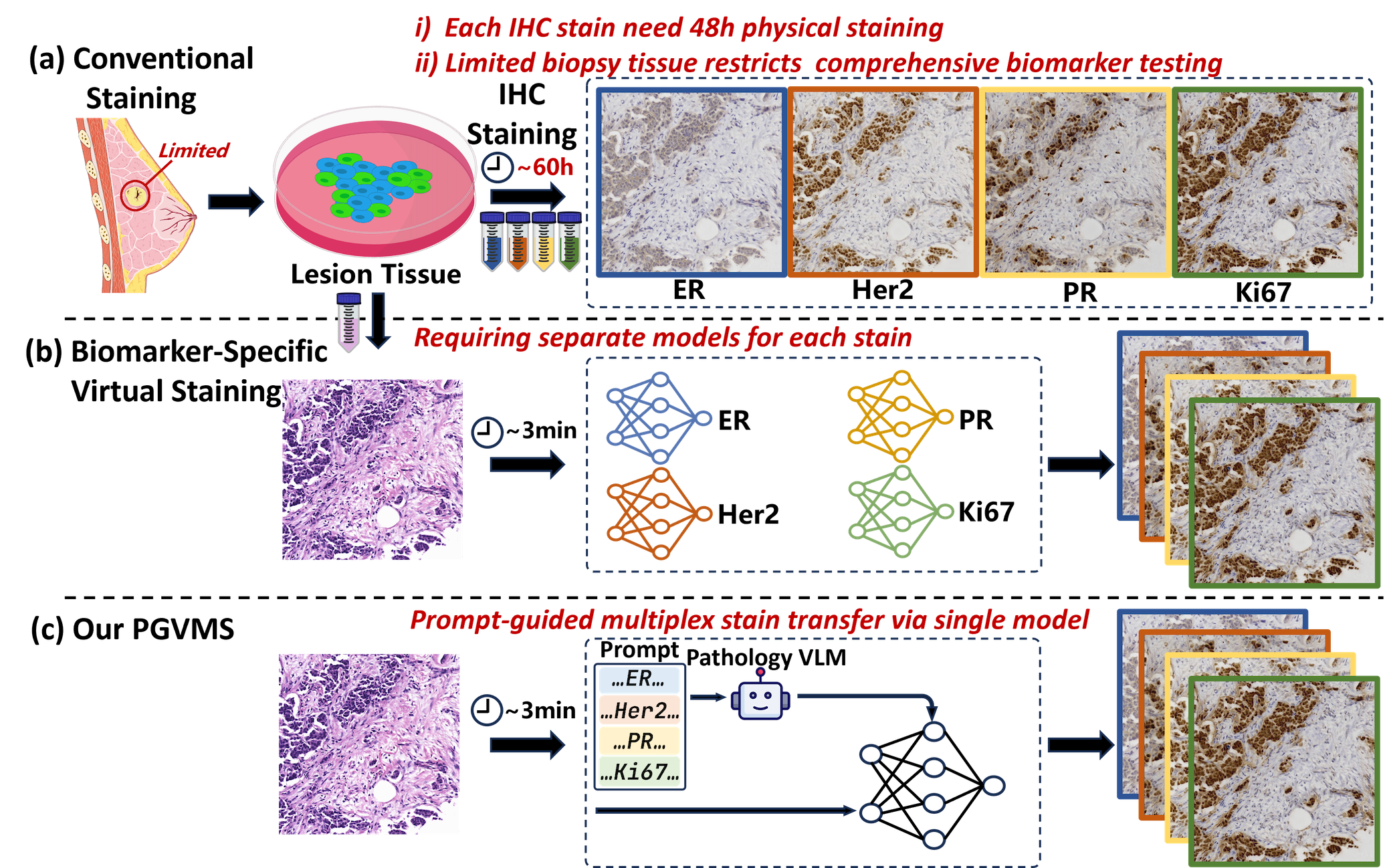}}
\caption{The comparison of traditional IHC staining and virtual staining workflows. From 
 (a) to (c), they are conventional IHC staining, biomarker-specific virtual staining and our PGVMS.}
\label{intro_1}
\end{figure}

\IEEEPARstart{C}{ancer} diagnosis primarily depends on histopathological analysis, with hematoxylin and eosin (H\&E) staining remaining the gold standard by providing exceptional visualization of tissue morphology through its characteristic nuclear (blue/purple) and cytoplasmic (pink) differentiation\cite{irshad2014automated}.
While H\&E staining offers excellent morphological assessment, molecular biomarker identification requires more advanced techniques such as multiplex immunohistochemistry (mIHC), which enables simultaneous detection of multiple critical biomarkers within a single tissue section \cite{weitz2023multi}.
However, mIHC analysis faces dual practical limitations: (1) insufficient tissue quantities in small biopsies constrain comprehensive biomarker assessment, and (2) specialized equipment and time-consuming protocols (up to 60 hours \cite{harms2023multiplex}) restricts its availability in resource-limited settings (Fig. \ref{intro_1}a).

\begin{figure}[t]
\centerline{\includegraphics[width=\columnwidth]{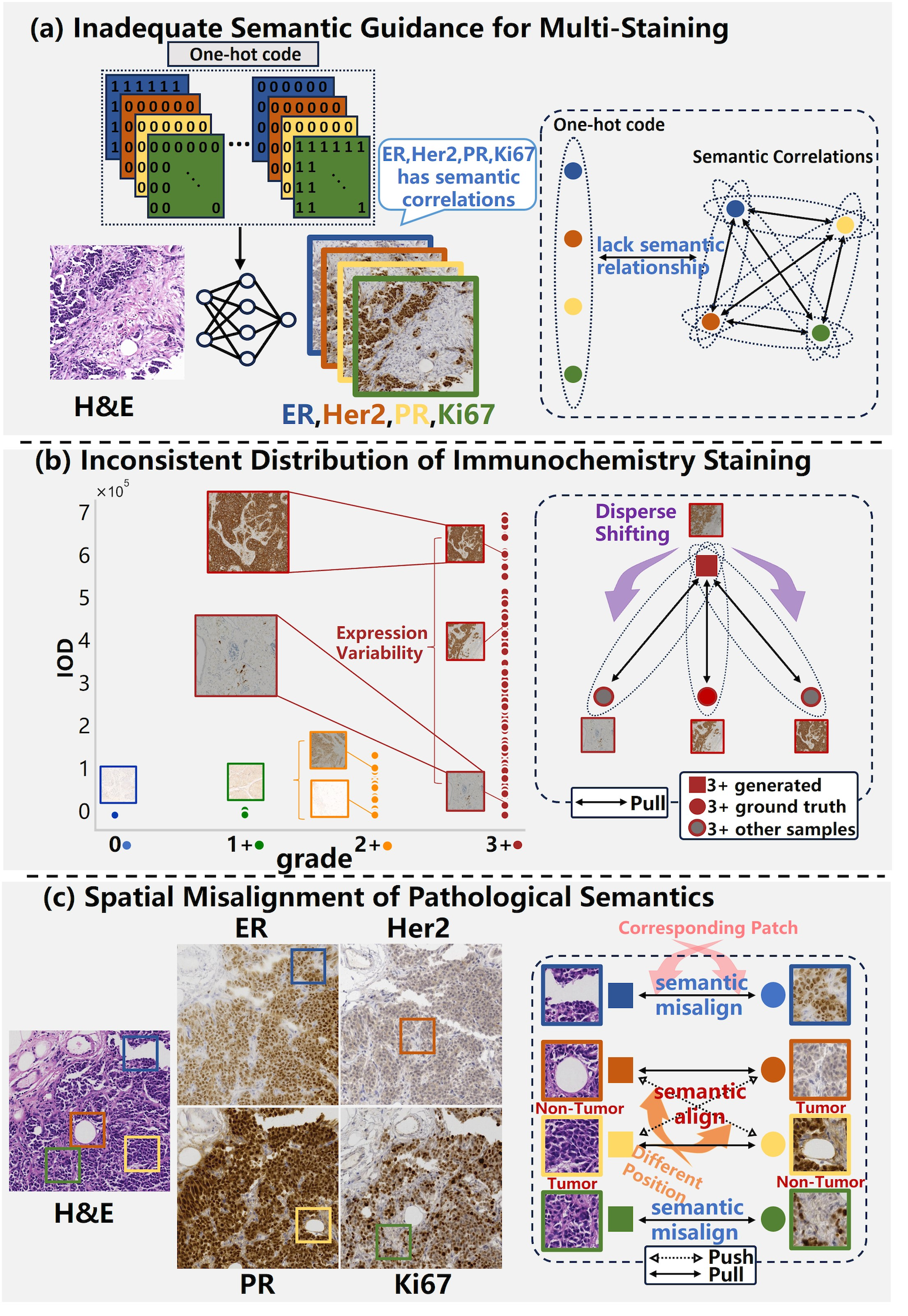}}
\caption{Three challenges of virtual multi-staining. (a) Inadequate semantic guidance for multi-staining. (b) Inconsistent distribution of immunochemistry staining. (c) Spatial misalignment of pathological semantics.}
\label{intro_2}
\end{figure}

Virtual staining has emerged as a promising computational approach for pathological analysis, as illustrated in Fig. \ref{intro_1}b. This AI-based technique enables digital transformation of H\&E-stained images into IHC representations, simulating biomarker expression patterns without physical staining. However, biomarker-specific virtual staining methods face inefficiency and inconsistency issues, as they require separate models for each stain.

The field of virtual staining has evolved through three stages: (1) initial GAN-based implementations lacking pathological constraints \cite{isola2017image,park2020contrastive,zhu2017unpaired}, (2) subsequently, intermediate solutions incorporating hierarchical semantic information at progressively finer scales (grade-level \cite{zeng2022semi}, patch-level \cite{liu2022bci,zhang2022mvfstain,li2023adaptive}), and (3) some solutions achieving pixel-level precision \cite{liu2021unpaired,peng2024advancing,chen2024pathological}.
Significantly, recent breakthroughs enable controlled multi-marker generation through various approaches: Guan et al. \cite{guan2025correlated} developed multi-stain using a shared encoder with four dedicated decoders to exploit inter-marker correlations, while other methods employ one-hot encoding for discrete biomarker selection \cite{lin2022unpaired} or CLIP-based text embedding for prompt control \cite{dubey2024vims}.

Rethinking the existing methods and challenges of virtual multiplex staining in H\&E to IHC, three significant challenges emerge (Fig. \ref{intro_2}): 
\textbf{(a) Inadequate semantic guidance for multi-staining}: Existing virtual multi-staining methods primarily employ one-hot encoding, which artificially enforces categorical independence among molecular markers while disregarding their inherent biological correlations in tissue expression. Although CLIP-based prompt guidance has been proposed to mitigate this limitation, its general-domain pretraining lacks pathology-specific adaptation, resulting in imprecise histopathological semantic capture. Consequently, current frameworks cannot effectively achieve semantic-aware co-staining generation for multiple biomarkers.
\textbf{(b) Inconsistent distribution of immunochemistry staining}: Current approaches fail to adequately capture fine-grained molecular semantics in pathological images, particularly the precise protein expression levels within specific stained regions. This limitation is especially evident in 3+-grade whole slide images (WSIs), where significant intra-grade variation in protein expression levels exists. By relying solely on coarse grade-level annotations rather than quantifying actual protein expression, existing methods disperse critical molecular-level pathological information, resulting in inconsistent immunochemistry staining distributions between generated and ground-truth IHC images. 
\textbf{(c) Spatial misalignment of pathological semantics}: Typically, virtual staining GT pairs are obtained from two depth-wise consecutive cuts of the same tissue, which are then stained separately. This approach inevitably prevents pixel-perfect image correspondences due to changes in cell morphology and staining-induced degradation. Inherent spatial inconsistencies exist between input images (H\&E) and label images (IHC), meaning that corresponding patches in paired images may contain different normal and tumor cells. Representations with similar semantics may be incorrectly separated and those with opposing semantics may be incorrectly merged. This affects the training efficiency of the model and the semantic accuracy of the generated images.

To address the aforementioned challenges, we propose a novel prompt-guided framework for virtual multiplex IHC staining (PGVMS). PGVMS advances virtual multiplex staining by incorporating semantic-preserving mechanisms to maintain critical pathological relationships across different staining modalities. We introduce one novel generator and two innovative learning strategies to tackle the corresponding problems:
(a) For inadequate semantic guidance on multi-stain generation, we propose a pathological semantics-style guided (PSSG) generator  that leverages the CONCH pathology visual language model for prompt-guided stain generation. This approach enables efficient virtual multi-staining by incorporating rich semantic embeddings, ensuring accurate and context-aware transformations from H\&E to multiple IHC stains, such as ER, PR, HER2, and Ki67.
(b) For inconsistent distribution of immunochemistry staining, we directly quantify the protein expression in each IHC-stained image by calculating the optical density in the DAB channel \cite{harms2023multiplex}. This approach preserves molecular-level pathological semantics. {Additionally, we propose a novel focal optical density (FOD) map that adaptively re-weights contributions from protein-expressing regions and normal tissue during quantification, providing a more accurate and robust representation of protein expression levels.}
(c) For spatial misalignment of pathological semantics, we assume that the pathological content of the generated image align with the label. To enhance pathological semantic interaction despite spatial inconsistencies, we facilitate the convergence of pathological features towards cross-corresponding prototypes in both the generated image and the label. This ensures semantic consistency even in the presence of spatial variations.

The main contributions of this paper are as follows:
\begin{itemize}
\item[•] \revision{We develop a pathological semantics–style guided (PSSG) generator by integrating a pathology visual language model, enabling prompt-driven semantic embedding for biologically informed multiplex IHC generation.}

\item[•] \revision{To preserve molecular-level pathological semantics, a protein-aware learning strategy (PALS) is introduced to explicitly quantify and constrain protein expression distributions via optical density modeling.}

\item[•] \revision{We further propose a prototype-consistent learning strategy (PCLS) to align high-protein expression prototypes with the label, effectively addressing spatial misalignment.}

\item[•] \revision{Comprehensive experiments on the MIST, IHC4BC, and cross-organ clinical datasets demonstrate that the proposed method achieves state-of-the-art virtual staining accuracy while enabling efficient multiplex IHC generation.}

\end{itemize}

\section{Related works}
\subsection{Image-to-Image Translation}
Image-to-image translation (I2I) aims to transform an input image from a source domain to a target domain while preserving content and transferring style. Early methods like pix2pix \cite{isola2017image} and CycleGAN \cite{zhu2017unpaired} used conditional GANs and cyclic consistency for supervised and unpaired translation, respectively. Park et al. \cite{park2020contrastive} introduced Contrastive Unpaired Translation (CUT), which leverages contrastive learning in a patch-based manner to maximize mutual information between input-output pairs. However, some argue that traditional I2I methods often fail to capture domain-specific details and are limited to one-to-one mappings. For example, Kazemi et al. \cite{kazemi2018unsupervised}  extended CycleGAN to enable one-to-many mappings by learning domain-specific codes, while Almahairi et al. \cite{cyclegan2018learning} proposed augmented CycleGAN for many-to-many multimodal translations. These advancements, with CUT as a key innovation, highlight the importance of balancing content preservation, style transfer, and output diversity in I2I tasks. But these methods are constrained in direct application due to their inability to explicitly preserve pathological semantics.
{In recent years, diffusion-based I2I frameworks have emerged as powerful alternatives, offering stronger generative priors and controllable synthesis. For instance, ControlNet \cite{zhang2023adding} introduces conditional control structures to guide diffusion models with external cues such as edges or segmentation maps, enabling fine-grained structural preservation. UNSB (Unpaired Noise Scheduling Bridge) \cite{kim2023unpaired} formulates I2I as a noise-bridging process between source and target domains, improving domain adaptation and stability. DDBM (Denoising Diffusion Bridge Model) \cite{zhou2023denoising} further extends this concept by explicitly learning domain-conditioned diffusion paths, facilitating faithful translation while maintaining semantic consistency. These diffusion-based paradigms significantly enhance fidelity and controllability, marking a new era for I2I applications, including virtual staining.}
 
\subsection{Virtual Staining}
Virtual staining using GANs has advanced significantly in recent years. Existing methods focus on extracting pathological semantic information from inconsistent ground truth pairs. For example, Liu et al. \cite{liu2022bci} proposed PyramidP2P, a pyramid pix2pix image generation method, but it requires pixel-level paired data, which is hard to obtain. At the image level, Zeng et al. \cite{zeng2022semi} 
 ensured virtual IHC accuracy with a Pos/Neg classifier. At the patch level, Li et al. \cite{li2023adaptive} introduced ASP, which uses patch-based contrastive learning to tolerate partial data misalignment. At the pixel level, Liu et al. \cite{liu2021unpaired} transformed H\&E to Ki67 using a pathology representation network with expert-annotated labels. Peng et al. \cite{peng2024advancing} extracted tumor information from labeled images using a nuclear density detector to align tumor locations. However, these methods struggle with severe spatial misalignment. To address this, Chen et al. \cite{chen2024pathological} proposed PSPStain, which detects tumor protein expression in patches and uses prototype learning for cross-spatial semantic alignment.
 For virtual multiplex staining, Lin et al. \cite{lin2022unpaired} used style-guided normalization with one-hot encoding to transform one staining type into multiple types. Xiong et al. \cite{xiong2025unpaired} optimizes visual prompts to control content and style. Zhang et al. \cite{zhang2022mvfstain} disentangled style and content to generate multiple staining types while predicting positive signals. Guan et al. \cite{guan2025correlated} developed Multi-IHC, using a shared encoder and four decoders. 
 Dubey et al. \cite{dubey2024vims} introduced VIMs, leveraging CLIP \cite{radford2021learning} for prompt embeddings (e.g., \revision{``CDX2"}) and fine-tuning a diffusion model with LoRA for prompt-controlled multi-virtual staining.



\revision{Despite their potential, current virtual multiplex staining methods face notable limitations. One-hot encoding approaches \cite{lin2022unpaired} treat molecular markers as independent categories, ignoring biological correlations and limiting semantic richness. Prompt-guided methods, such as Dubey et al. \cite{dubey2024vims}, employ CLIP embeddings with LoRA-finetuned diffusion models for prompt-controlled multiplex staining. However, CLIP is pretrained on general-domain text-image data rather than pathology-specific pairs, so it fails to capture the semantic relationships among different IHC stains. Other methods, such as Zhang et al. \cite{zhang2022mvfstain}, cannot control which stain is generated individually, producing multiple stains simultaneously. Additional approaches \cite{liu2022bci,zeng2022semi,li2023adaptive,liu2021unpaired,peng2024advancing,chen2024pathological,guan2025correlated} rely on biomarker-specific models, which are computationally inefficient and do not scale well for multiple stains. These limitations motivate the need for pathology-aware semantic guidance mechanisms to achieve efficient and biologically consistent multiplex virtual staining.}

\subsection{Pathological Visual Language Model}
In recent years, the development of pathological visual language models (VLMs) has significantly advanced computational pathology, enabling more accurate and efficient analysis of histopathological images. {Among these models, MUSK \cite{xiang2025vision}, PLIP \cite{huang2023visual}, and CONCH \cite{lu2024visual} stand out as notable contributions. MUSK, a vision-language foundation model for precision oncology, achieves state-of-the-art performance by employing a two-stage training paradigm: first, masked learning on 50 million tissue patches and 1 billion textual tokens, followed by contrastive learning on 1 million image-text pairs.} PLIP, on the other hand, utilizes a dataset of over 200,000 pathology image-text pairs sourced from medical Twitter, which enables PLIP to perform tasks like zero-shot classification and cross-modal retrieval. CONCH \cite{lu2024visual} focus on IHC image-text pairs, trained on 1.17 million pathology image-caption pairs. It excels in generating representations for non-H\&E stains, such as IHC and special stains, capturing nuanced relationships between images and text. These models significantly advance computational pathology through large-scale data and multi-modal integration.

\begin{figure*}[!htbp]
\centerline{\includegraphics[width=\textwidth]{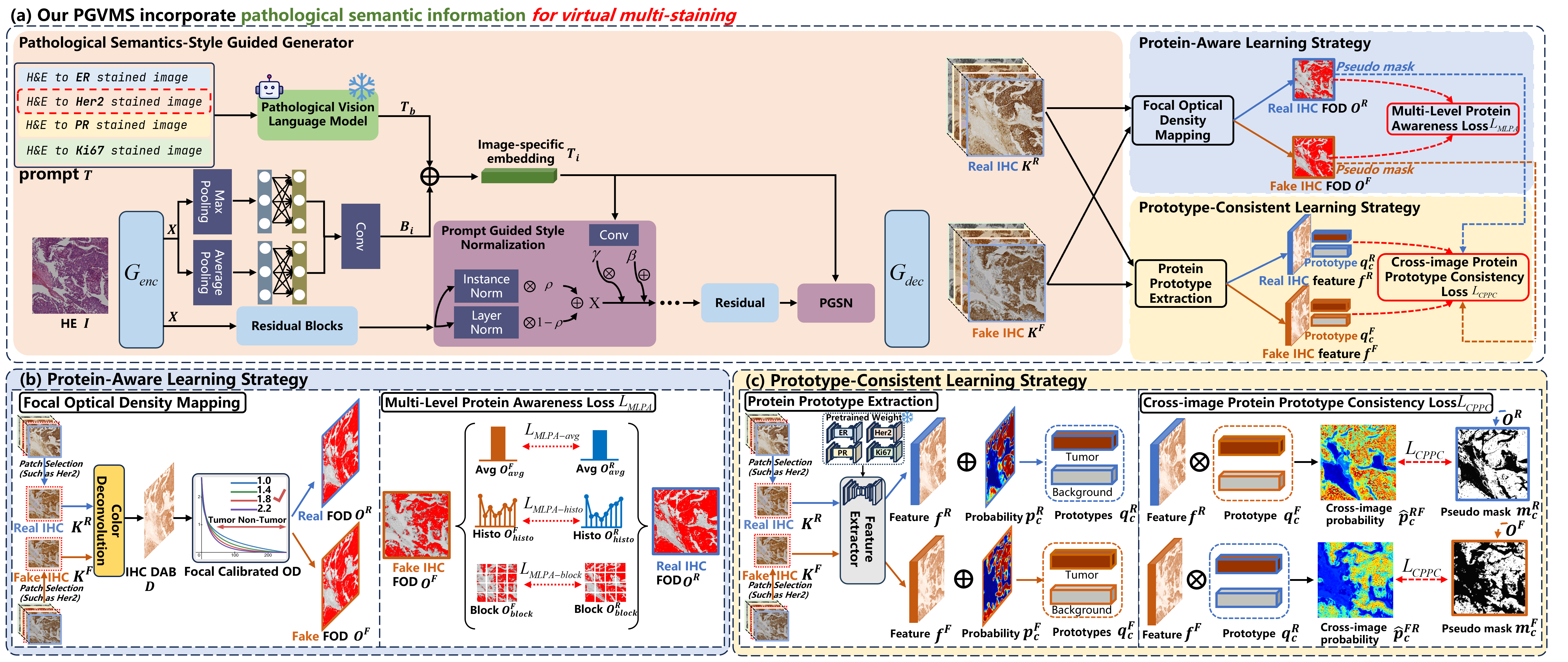}}
\caption{\revision{The framework of PGVMS. 
(a) Overall architecture of PGVMS, including the pathological semantics--style guided (PSSG) generator and schematic overviews of the protein-aware learning strategy (PALS) and the prototype-consistent learning strategy (PCLS). (b) and (c) show detailed illustrations of PALS and PCLS in the bottom row, highlighting their key components and operations.
}}
\label{method_1}
\end{figure*}

\section{Method}

This section presents the proposed PGVMS framework for generate mIHC images from H\&E images, as shown in Fig. \ref{method_1}. The PGVMS framework comprises three key components designed to achieve optimal virtual staining performance. First, we introduce a pathological semantics-style guided (PSSG) generator that employs prompt-based control to enable efficient multi-stain generation within a single unified model. To ensure pathological fidelity, the framework incorporates two specialized learning strategies: (1) protein-aware learning strategy (PALS) that maintains protein expression consistency across staining modalities, and {(2) prototype-consistent learning strategy (PCLS) that enhances high-protein expression region alignment with established pathological prototypes. }Together, these strategies guarantee the preservation of critical pathological features between virtually generated and physically stained IHC images.


 \subsection{Pathological Semantics-Style Guided Generator}




Our framework enables efficient multi-stain generation through semantic prompt control, leveraging CONCH's pathology-specific embeddings trained on 1.17 million IHC image-text pairs \cite{lu2024visual}. Given text prompt $T$ (e.g., \revision{``H\&E to Her2 stained image"}), we first extract its base representation:


\begin{equation}
    T_b = \text{CONCH}(T)
\end{equation}
where $T$ represents the natural language prompt and $T_b$ captures its general IHC-related semantics. This foundation model embedding provides crucial prior knowledge about biomarker expression patterns and their pathological significance.

To better align embeddings with individual tumor characteristics, we develop an adaptive image-conditioned bias modulation approach. The H\&E image features $X = G_{enc}(I)$ are processed through dual two pooling branches:

\begin{equation}
    B_i = \text{Conv}(\text{Cat}(\text{MLP}(\text{AvgPool}(X)),\text{MLP}(\text{MaxPool}(X))))
\end{equation}

The average pooling branch captures global tissue architecture, while max pooling emphasizes local cellular abnormalities. Their concatenation and subsequent convolution yield a tumor-specific bias term $B_i$ that refines the prompt embedding:

\begin{equation}
    T_i = T_b + B_i
\end{equation}

{This adjustment ensures the staining process adapts to both the semantic prompt and the unique morphological context of each high-protein expression region.}

The prompt guided style normalization (PGSN) module then modulates the staining process through:
\begin{equation}
    \text{PGSN}(X) = \gamma \cdot (\rho\cdot\text{IN}(X) + (1-\rho)\cdot\text{LN}(X)) + \beta
\end{equation}
where $\gamma,\beta = \text{Conv}(T_i)$ transform semantic embeddings into stain-specific parameters, and $\rho$ automatically balances instance normalization (preserving local staining patterns) against layer normalization (maintaining global tissue structure). This adaptive mixture respects both prompt-derived staining styles and inherent tissue organization.


After $N$ alternating residual and PGSN blocks, the decoder generates the target stain:
\begin{equation}
    K^F = G_{dec}(\underbrace{\text{PGSN}(\text{Res}(\cdots\text{PGSN}(\text{Res}(X))))}_{N \text{ layers}})
\end{equation}

The complete architecture achieves two key innovations: {First, it bridges general pathological knowledge from foundation models with case-specific high-protein expression characteristics through adaptive prompt tuning. } Second, the hybrid normalization strategy respects both semantic staining directives and biological tissue constraints. 

\subsection{Protein-Aware Learning Strategy}
In IHC staining, target proteins are typically detected through enzyme-conjugated antibodies, with horseradish peroxidase being commonly employed \cite{harms2023multiplex}.
DAB chromogen forms a brown precipitate with peroxidase, allowing protein quantification via DAB channel optical density. Based on these and Chen et al. \cite{chen2024pathological} , PALS initially employs focal optical density mapping to quantify protein distribution in both generated and ground truth images, then applys multi-level protein awareness to enforce biochemical constraints across three hierarchical levels.

\subsubsection{Focal Optical Density Mapping}

Firstly, $K^R$ and $K^F$ undergo subsequent processing to get the FOD.
The optical density (OD) provides a quantitative measure of stain concentration according to the Lambert-Beer law \cite{varghese2014ihc}:
\begin{equation}\label{OD}
    OD_C=-log_{10}(K_C/K_{0,C})=A \ast \boldsymbol{c}_C
\end{equation}
where $K_{0,C}$ and $K_C$ represent the incident and transmitted light intensities respectively for channel $C$, $A$ denotes the stain amount, and $\boldsymbol{c}_C$ is the absorption factor.

Traditional color deconvolution enables DAB stain separation. The stain concentrations are obtained through:

\begin{equation}
{S} = {M}^{-1}\cdot{OD}
\end{equation}
where ${M} \in {R}^{3\times 3}$ is the stain matrix containing absorption coefficients, ${S} \in {R}^{3}$ represents stain concentration vectors. 

Only selecting DAB channel transform to standard OD:

\begin{equation}
D = {M}_{DAB} \cdot {S}_{DAB}
\end{equation}

However, the typically small {protein expression regions relative to normal tissue create a severe class imbalance. To address this, we introduce a focal optical density (FOD) transformation that emphasizes protein expression regions while suppressing normal tissue}:
\begin{equation}
    {O_C}=(-log_{10}((D_C)/D_{0,C}))^{\alpha}
\end{equation}
where $\alpha > 1$ is a tunable focusing parameter. As shown in Fig.~\ref{method_1}(b), increasing $\alpha$ enhances {protein expression region }emphasis, with $\alpha=1.8$ providing optimal performance in our experiments.

\subsubsection{Multi-Level Protein Awareness}
The multi-level protein awareness (MLPA) loss enforces protein expression consistency between generated ($O^F$) and real ($O^R$) IHC images through a hierarchical constraint framework operating at three biologically meaningful spatial scales:

\textbf{Global Protein Expression Constraint:} 
At the macroscopic level, we constrain the overall staining intensity while allowing for natural biological variations:
\begin{equation}
L_{MLPA-{avg}} =  
\begin{cases}
\left\|\Delta O\right\|_{2}, & \text{if } |\Delta O| \geq \beta \cdot O_{avg}^R \\
0, & \text{if } |\Delta O| < \beta \cdot O_{avg}^R
\end{cases}
\end{equation}
where $\Delta O = O_{avg}^F -O_{avg}^R$ quantifies the mean intensity difference between generated and real images, and $\beta=0.2$ provides a 20\% tolerance threshold that accommodates acceptable clinical variations in staining intensity while preserving diagnostically significant differences.

\textbf{Histogram Distribution Alignment:} The histogram term ensures proper representation of all protein expression levels by comparing normalized intensity distributions across $N_h=20$ bins:
\begin{equation}
     L_{MLPA-{histo}} = \frac{1}{N_h}\sum_{i=1}^{N_h} \left\|O_{histo_i}^F-O_{histo_i}^R\right\|_{2}
\end{equation}
This constraint operates on the complete dynamic range of staining intensities, maintaining proportional representation of clinically relevant expression levels including weak positive regions (0 and 1+), moderate expression (2+), and strong overexpression (3+). The bin-wise comparison prevents common artifacts such as intensity clustering or unrealistic expression distributions.

\textbf{Regional Pattern Consistency:} The regional consistency term preserves local biomarker expression patterns through $N_b=16$ non-overlapping image blocks:
\begin{equation}
     L_{MLPA-{block}} = \frac{1}{N_b}\sum_{i=1}^{N_b} \left\|O_{block_i}^F-O_{block_i}^R\right\|_{2}
\end{equation}
{This spatial constraint ensures proper maintenance of protein expression heterogeneity patterns}, accurate localization of biomarker hotspots, and prevention of unrealistic expression mosaicism. The block-wise comparison mimics pathologists' regional analysis workflow during diagnostic evaluation.

The complete MLPA loss integrates these components:
\begin{equation}
    L_{MLPA} = L_{MLPA-{avg}} + L_{MLPA-{histo}} + L_{MLPA-{block}}
\end{equation}

\subsection{Prototype-Consistent Learning Strategy}

{To address the challenge of spatial misalignment between generated and reference IHC images, we propose PCLS, a method grounded in the biological principle that protein expression content remains intrinsically consistent across different staining modalities of the adjacent patches. The PCLS first extracts protein prototypes and features from both generated and reference images through protein prototype extraction, then enforces bidirectional cross-image protein prototype consistency.}
\subsubsection{Protein Prototype Extraction}
For each IHC stain type, the prototype extraction process utilizes a corresponding frozen pretrained segmentation UNet to analyze both generated ($K^F$) and real ($K^R$) IHC images through three sequential operations.

First, the UNet encoder-decoder architecture processes each image to extracts high-dimensional feature maps $f^F, f^R \in \mathbb{R}^{H\times W \times D}$ from its penultimate layer:
\begin{equation}
f^F, f^R  = \text{UNet}_{\text{penultimate}}(K^F,K^R)
\end{equation}

{
Second, the final UNet layer outputs pixel-wise classification probabilities $p_c^F, p_c^R$ through softmax activation, quantifying protein expression ($c=1$) versus normal ($c=0$) tissue likelihoods.}
\begin{equation}
p_c^F, p_c^R = \text{softmax}(\text{UNet}_{\text{final}}(K^F,K^R))
\end{equation}
where $c \in \{\text{protein expression region}, \text{ normal tissue}\}$ indicates the tissue class.

Third, the prototype computation applies confidence-aware feature aggregation by combining the probability maps with their corresponding features:
\begin{align}
    {q}_c^{{F}}=\frac{\sum_i p_c^F(i)\cdot f^F(i)}{\sum_i p_c^F(i)} ,
    & {q}_c^{{R}}=\frac{\sum_i p_c^R(i)\cdot f^R(i)}{\sum_i p_c^R(i)} 
\end{align}

\subsubsection{Cross-image Protein Prototype Consistency}
The cross-image consistency mechanism enforces bidirectional semantic alignment through cosine similarity measures:
\begin{align}
    \hat{s}_{c}^{FR}=\frac{f^F\cdot{q}_c^R}{\|f^F\|\cdot\left\|{q}_c^R\right\|},
    & \hat{s}_{c}^{RF}=\frac{f^R\cdot{q}_c^F}{\|f^R\|\cdot\left\|{q}_c^F\right\|}
\end{align}

These similarity metrics are converted to probability distributions via softmax normalization:
\begin{align}
    \hat{p}_c^{FR} =\frac{e^{\hat{s}_{c}^{FR}}}{\sum_{c}e^{\hat{s}_{c}^{FR}}},
    & \hat{p}_c^{RF} = \frac{e^{\hat{s}_{c}^{RF}}}{\sum_{c}e^{\hat{s}_{c}^{RF}}}
\end{align}

{The complete cross-protein prototype consistency (CPPC) loss integrates these bidirectional constraints}:
\begin{equation}
L_{CPPC} = \frac{1}{C \times H \times W} \sum_{i=1}^{H \times W} \sum_{c=1}^{C} \left( L_{FR}(i, c) + L_{RF}(i, c) \right)
\end{equation}
where the individual loss components measure the deviation between predicted and actual protein regions:
\begin{align*}
L_{FR}(i, c) &= \left\| \hat{p}_c^{FR}(i) - m_c^F(i) \right\|_{2} \\
L_{RF}(i, c) &= \left\| \hat{p}_c^{RF}(i) - m_c^R(i) \right\|_{2}
\end{align*}

\noindent where $m_c^F(i)$ and $ m_c^R(i)$ are derived from $O^F$ and $O^R$, respectively. This formulation simultaneously ensures: (1) cross-modality protein morphology consistency, (2) robustness to staining variations, (3) stain-invariant feature learning, and (4) balanced bidirectional constraint enforcement.

\subsection{Other Loss Function}
{
The adversarial loss encourages the generated image $K^F$ to be indistinguishable from the real image $K^R$. It is formulated as:}
{
\begin{equation}
{L}_{{adv}} = 
\mathbb{E}_{K^R} \big[\log D(K^R) \big] +
\mathbb{E}_{K^F} \big[\log (1 - D(K^F)) \big],
\end{equation}
}
{where $D(\cdot)$ denotes the discriminator, $K^F$ is the generated image (fake), and $K^R$ is the real image (ground truth). This loss encourages the generator to produce outputs that are realistic and indistinguishable from real samples.}

{
The NCE loss~\cite{oord2018representation,park2020contrastive} aims to preserve content consistency across image translation by maximizing mutual information between corresponding patches in the input and output images. It is formulated as a patch-based InfoNCE objective~\cite{oord2018representation}:}

{
\begin{equation}
{L}_{{NCE}} = - \log \frac{\exp(\hat{z}_Y \cdot z_X / \tau)}
{\exp(\hat{z}_Y \cdot z_X / \tau) + \sum_{n=1}^{N} \exp(\hat{z}_Y \cdot \hat{z}_X^n / \tau)},
\end{equation}
}
{where $\hat{z}_Y$ is the embedding of a patch in the output image (anchor), $z_X$ is the corresponding input patch embedding (positive), $\hat{z}_X^n$ are embeddings of non-corresponding patches (negatives), and $\tau$ is a temperature hyperparameter controlling the sharpness of the softmax distribution.}

{The structural similarity (SSIM) loss encourages the generated image $K^F$ to preserve the structural information of the real image $K^R$:}
{
\begin{equation}
{L}_{{SSIM}} = 1 - \text{SSIM}(K^F, K^R),
\end{equation}
}
{where $\text{SSIM}(\cdot, \cdot)$ computes the structural similarity index between two images. Minimizing ${L}_{{SSIM}}$ promotes preservation of luminance, contrast, and structural patterns in the generated image.}

{
The gaussian pyramid (GP) loss \cite{liu2022bci} enhances multi-scale consistency between generated and ground truth images, relaxing the strong constraints of the standard L1 loss.}

{The loss at scale $i$ is defined as:}
{
\begin{equation}
S_i = \mathbb{E}_{x,y,z}\left[\|Gaussian_i(K^R) - Gaussian_i(K^F))\|_1\right]
\end{equation}
}
{where $Gaussian_i$ denotes the Gaussian filtering operation at scale $i$, $K^F$ is the output image, $K^R$ is the ground truth.}

{The overall multi-scale loss is formulated as a weighted sum across scales:}
{
\begin{equation}
L_{{GP}} = \sum_{i} \lambda_i S_i
\end{equation}
}
{where $\lambda_i$ represents the weight for scale $i$.}

\subsection{Overall Loss Function}
The overall learning objective is as follows:
\begin{equation}
\begin{aligned}
L_{total} &= L_{adv} + L_{NCE} + \lambda_{M}L_{MLPA} \\
&\quad + \lambda_{C}L_{CPPC} + \lambda_{S}L_{SSIM} + \lambda_{G}L_{GP}
\end{aligned}
\end{equation}
where $L_{MLPA}$, $L_{CPPC}$, $L_{GP}$ focus on pathological consistency, with $L_{GP}$ originating from \cite{liu2022bci}. $L_{adv}, L_{NCE}$, $L_{SSIM}$ contribute to image quality enhancement.

\section{ EXPERIMENTS AND RESULTS}

Our method achieves superior performance on both MIST and IHC4BC datasets for H\&E to IHC multi-stain transfer, enabling efficient prompt-guided virtual staining of multiple biomarkers.
The comparative experimental results show that our method achieve the virtual multiple staining with prompt guided and outperforms other existing methods.
\subsection{Datasets}
\subsubsection{MIST dataset} The MIST dataset~\cite{li2023adaptive} provides precisely aligned H\&E-IHC image pairs for four critical breast cancer biomarkers (Ki67, ER, PR, and HER2). All image patches are non-overlapping 1024×1024 pixels, with substantial training pairs per stain (4,161-4,642) from 56-64 whole slide images (WSIs), each accompanied by a standardized test set of 1,000 image pairs.
\subsubsection{IHC4BC dataset} The IHC4BC dataset~\cite{akbarnejad2023predicting} offers larger-scale aligned H\&E-IHC pairs for the same four biomarkers, containing approximately 90,000 total patches at 1000×1000 resolution. With 52-60 WSIs per stain, it provides significantly more training data (16,995-26,395 pairs per biomarker) while maintaining the same 1,000-pair test set standardization. 

Both datasets maintain overall structural alignment despite minor local deformations, serving as comprehensive benchmarks for computational pathology research.

\begin{table*}[ht]
  \centering
  \setlength\tabcolsep{4.5pt}
  \renewcommand{\arraystretch}{1}
  \caption{\revision{COMPARISON OF VIRTUAL STAINING PERFORMANCE ON \textbf{MIST} DATASET. 
  THE BEST SCORES ARE IN BOLD. 
  MODELS ARE GROUPED INTO ONE-TO-ONE AND ONE-TO-MANY CATEGORIES.
}}
  \label{tab:all_datasets}

  \resizebox{\textwidth}{!}{
    \begin{tabular}{c|c|cc|cc|cc|cc|cc|cc}
      \Xhline{1.5pt}
      \multirow{3}{*}{Type} & 
      \multirow{3}{*}{Method} & 
      \multicolumn{6}{c|}{HER2} & \multicolumn{6}{c}{ER} \\
      \cline{3-14}
      & & \multicolumn{2}{c|}{Quality (reference)} & \multicolumn{2}{c|}{Pathology} & \multicolumn{2}{c|}{Perception} & 
      \multicolumn{2}{c|}{Quality (reference)} & \multicolumn{2}{c|}{Pathology} & \multicolumn{2}{c}{Perception} \\
      \cline{3-14}
      & & PSNR$\uparrow$ & SSIM$\uparrow$ & IOD$_{\times10^7}$ & Pearson-R$\uparrow$ & FID$\downarrow$ & DISTS$\downarrow$ &
      PSNR$\uparrow$ & SSIM$\uparrow$ & IOD$_{\times10^7}$ & Pearson-R$\uparrow$ & FID$\downarrow$ & DISTS$\downarrow$ \\
      \hline
      \multirow{9}{*}{One-to-one} 
      & CycleGAN~\cite{zhu2017unpaired} & 
      12.2042 & 0.1588 & +4.5848 & 0.1995 & 63.8952 & 0.2864 & 
      13.2516 & 0.2187 & -1.6250 & 0.2916 & 54.1309 & 0.2730 \\
      
      & Pix2Pix~\cite{isola2017image} & 
      12.9840 & 0.1676 & -2.5782 & 0.2986 & 84.3667 & 0.2699 & 
      \textbf{14.7293} & 0.2200 & -5.8452 & 0.6307 & 90.6261 & 0.3050 \\
      
      & CUT~\cite{park2020contrastive} & 
      13.8998 & 0.1680 & -2.9478 & 0.7164 & 53.2911 & 0.2641 & 
      14.2172 & 0.2201 & -4.5073 & 0.5841 & 45.8074 & 0.2630 \\
      
      & ASP~\cite{li2023adaptive} & 
      14.1841 & 0.2004 & -5.7422 & 0.0659 & 50.8364 & 0.2517 & 
      14.2088 & 0.2178 & -4.5875 & 0.4802 & 54.7745 & 0.2705 \\
      
      & PyramidP2P~\cite{liu2022bci} & 
      \textbf{14.9122} & 0.1995 & -4.5285 & 0.6894 & 91.3455 & 0.2628 & 
      14.2062 & 0.2023 & -4.2775 & 0.7418 & 92.9292 & 0.2703 \\
      
      & TDKStain~\cite{peng2024advancing} &
      14.7620 & 0.1901 & -3.1244 & 0.8147 & 62.8442 & 0.2515 & 
      14.3451 & 0.2113 & -2.6321 & 0.8466 & 46.7817 & 0.2483 \\
      
      & PSPStain~\cite{chen2024pathological} & 
      14.1948 & 0.1876 & -2.5491 & 0.8303 & \textbf{41.3439} & 0.2541 & 
      14.5718 & 0.2143 & -2.0571 & 0.8763 & 39.8844 & 0.2431 \\

      & DDBM~\cite{zhou2023denoising} &
      14.7785 & 0.2156 & -6.4420 & -0.0228 & 201.5442 & 0.3426 &
      14.6963 & 0.2368 & -7.6428 & 0.1280 & 191.6672 & 0.3336 \\
      
      & UNSB~\cite{kim2023unpaired} &
      13.2797 & 0.2121 & -1.8809 & 0.3141 & 54.6514 & 0.3015 &
      13.3957 & 0.2452 & -1.5399 & 0.3981 & 39.1415 & 0.2955 \\

      \Xhline{0.8pt}
      \rowcolor{gray!10}
    & \cellcolor{gray!10}ControlNet~\cite{zhang2023adding} &
    8.6754 & 0.1658 & +3.2796 & 0.0080 & 121.8161 & 0.3882 &
    8.4172 & 0.1564 & +19.2318 & 0.0505 & 155.9284 & 0.3499 \\
    \rowcolor{gray!10}
    & \cellcolor{gray!10}VIMs~\cite{dubey2024vims} & 
    13.9114 & 0.1953 & -7.0886 & 0.3049 & 130.5218 & 0.3077 &
    14.1536 & 0.1969 & -2.9966 & 0.0434 & 97.3877 & 0.2979 \\
    \rowcolor{gray!10}
    & \cellcolor{gray!10}UMDST~\cite{lin2022unpaired} & 
    12.6934 & \textbf{0.2181} & 1.7366 & 0.1037 & 71.4340 & 0.3124 &
    13.6387 & \textbf{0.2528} & -1.5309 & 0.4156 & 59.9249 & 0.3122 \\
   
    \rowcolor{gray!10}
   \multirow{-4}{*}{One-to-many}  & \cellcolor{gray!10}PGVMS & 
    13.9246 & 0.1769 & \textbf{-0.4390} & \textbf{0.8548} & 44.4247 & \textbf{0.2336} &
    14.3583 & 0.2072 & \textbf{-0.5193} & \textbf{0.8902} & \textbf{36.5734} & \textbf{0.2299} \\

      \hline
      \multirow{3}{*}{Type} & 
      \multirow{3}{*}{Method} & 
      \multicolumn{6}{c|}{PR} & \multicolumn{6}{c}{Ki67} \\
      \cline{3-14}
      & & \multicolumn{2}{c|}{Quality (reference)} & \multicolumn{2}{c|}{Pathology} & \multicolumn{2}{c|}{Perception} & 
      \multicolumn{2}{c|}{Quality (reference)} & \multicolumn{2}{c|}{Pathology} & \multicolumn{2}{c}{Perception} \\
      \cline{3-14}
      & & PSNR$\uparrow$ & SSIM$\uparrow$ & IOD$_{\times10^7}$ & Pearson-R$\uparrow$ & FID$\downarrow$  & DISTS$\downarrow$ 
      & PSNR$\uparrow$ & SSIM$\uparrow$ & IOD$_{\times10^7}$ & Pearson-R$\uparrow$ & FID$\downarrow$ & DISTS$\downarrow$ \\
      \hline
      
      \multirow{9}{*}{One-to-one} 
      & CycleGAN~\cite{zhu2017unpaired} & 
      13.8537 & 0.2040 & -6.1662 & 0.2229 & 75.4442 & 0.2736 & 
      13.9144 & 0.2186 & -2.6220 & 0.2726 & 46.1386 & 0.2538 \\
      
      & Pix2Pix~\cite{isola2017image} & 
      \textbf{15.0557} & 0.2338 & -5.8785 & 0.7201 & 80.4802 & 0.3101 & 
      14.6090 & 0.2360 & -2.0715 & 0.5560 & 93.7772 & 0.2568 \\
      
      & CUT~\cite{park2020contrastive} & 
      14.3480 & 0.2209 & -2.8106 & 0.7032 & 41.2165 & 0.2601 & 
      14.4081 & 0.2095 & -2.2291 & 0.4878 & 42.0596 & 0.2520 \\
      
      & ASP~\cite{li2023adaptive} & 
      14.1200 & 0.2111 & -5.7499 & 0.5557 & 47.8368 & 0.2547 & 
      14.5852 & 0.2286 & -2.3507 & 0.4562 & 50.9120 & 0.2462 \\
      
      & PyramidP2P~\cite{liu2022bci} & 
      14.2600 & 0.2097 & -5.9274 & 0.7318 & 85.6816 & 0.2718 & 
      13.9477 & 0.2231 & -0.6542 & 0.5877 & 87.8460 & \textbf{0.2417} \\

      & TDKStain~\cite{peng2024advancing} &
      14.7783 & 0.2320 & -2.5365 & 0.8658 & 56.2217 & 0.2534 &
      14.6585 & 0.2391 & -1.0361 & 0.7473 & 58.5757 & 0.2454 \\
      
      & PSPStain~\cite{chen2024pathological} & 
      14.8824 & 0.2347 & -3.5771 & 0.8661 & 38.3390 & 0.2454 & 
      14.2602 & 0.2415 & \textbf{-0.0671} & 0.7718 & 37.5574 & 0.2607 \\
      
      & DDBM~\cite{zhou2023denoising} &
      13.4919 & 0.2242 & -6.4651 & 0.4206 & 168.8190 & 0.3087 &
      14.1996 & 0.2357 & -1.8336 & 0.4149 & 158.1006 & 0.2909 \\
      
      & UNSB~\cite{kim2023unpaired} &
      13.7457 & \textbf{0.2572} & -1.2455 & 0.3906 & 41.3750 & 0.3136 &
      14.1265 & 0.2722 & -0.3999 & 0.3981 & \textbf{33.2999} & 0.2836 \\
      
       \Xhline{0.8pt}
      \rowcolor{gray!10}
      
      & ControlNet~\cite{zhang2023adding} &
      9.0790 & 0.1508 & +5.6869 & 0.0928 & 136.1356 & 0.3710 &
      10.2877 & \textbf{0.2831} & +6.4306 & 0.2751 & 292.4799 & 0.3955 \\
       \rowcolor{gray!10}
      & VIMs~\cite{dubey2024vims} & 
      13.3384 & 0.1591 & -4.2534 & 0.0341 & 115.3529 & 0.3069 &
      13.7166 & 0.1906 & -6.0985 & 0.2202 & 110.7464 & 0.3002 \\
     \rowcolor{gray!10}
      & UMDST~\cite{lin2022unpaired} & 
      13.7066 & 0.2515 & -1.7874 & 0.4968 & 59.4099 & 0.3039 &
      13.8163 & 0.2595 & +0.7868 & 0.2642 & 94.9168 & 0.3174 \\
     
       \rowcolor{gray!10}
      \multirow{-4}{*}{One-to-many}& PGVMS & 
      14.2510 & 0.2000 & \textbf{+0.7815} & \textbf{0.9126} & \textbf{34.9099} & \textbf{0.2300} & 
      \textbf{14.8344} & 0.2285 & -1.1125 & \textbf{0.8018} & 36.3537 & 0.2439 
      \\
      \Xhline{1.5pt} 
    \end{tabular}
  }
\end{table*}

\begin{table*}[ht]
  \centering
  \setlength\tabcolsep{5pt}
  \renewcommand{\arraystretch}{1}
  \caption{\revision{COMPARISON OF VIRTUAL STAINING PERFORMANCE ON \textbf{IHC4BC} DATASET. THE BEST SCORES ARE IN BOLD. MODELS ARE GROUPED INTO ONE-TO-ONE AND ONE-TO-MANY CATEGORIES.
}}
  \label{tab:ihc4bc_results}

  \resizebox{\textwidth}{!}{
    \begin{tabular}{c|c|cc|cc|cc|cc|cc|cc}
      \Xhline{1.5pt} 
       \multirow{3}{*}{Type} &\multirow{3}{*}{Method} & 
      \multicolumn{6}{c|}{HER2} & \multicolumn{6}{c}{ER} \\
      \cline{3-14}
      && \multicolumn{2}{c|}{Quality (reference)} & \multicolumn{2}{c|}{Pathology} & \multicolumn{2}{c|}{Perception} & 
      \multicolumn{2}{c|}{Quality (reference)} & \multicolumn{2}{c|}{Pathology} & \multicolumn{2}{c}{Perception} \\
      \cline{3-14}
      && PSNR$\uparrow$ & SSIM$\uparrow$ & IOD$_{\times10^8}$ & Pearson-R$\uparrow$ & FID$\downarrow$ & DISTS$\downarrow$ & PSNR$\uparrow$ & SSIM$\uparrow$ & IOD$_{\times10^7}$ & Pearson-R$\uparrow$ & FID$\downarrow$ & DISTS$\downarrow$ \\
      \hline
     \multirow{9}{*}{One-to-one} 
      &CycleGAN~\cite{zhu2017unpaired} & 10.1229 & 0.0930 & +1.9836 & 0.3071 & 143.8269 & 0.3385 & 17.3148 & 0.4406 & +0.8744 & 0.3878 & 50.9366 & 0.2545 \\
      &Pix2Pix~\cite{isola2017image} & 11.3910 & 0.1228 & -0.9739 & 0.6214 & 117.3533 & 0.3027 & 20.7537 & 0.4931 & -1.0956 & 0.2171 & 129.9073 & 0.3050 \\
      &CUT~\cite{park2020contrastive} & 10.9131 & 0.1227 & -1.2030 & -0.0613 & 63.3108 & 0.2657 & 19.4466 & 0.4745 & -1.0011 & 0.0705 & 41.8358 & 0.2618 \\
      &ASP~\cite{li2023adaptive} & 11.2034 & 0.1245 & -1.1742 & -0.0988 & 63.9972 & 0.2658 & 19.5277 & 0.4834 & -0.6886 & 0.1810 & 37.9753 & 0.2644 \\
      &PyramidP2P~\cite{liu2022bci} & 10.8870 & 0.1325 & -0.5059 & 0.4234 & 113.6700 & 0.3612 & \textbf{21.2933} & \textbf{0.5606} & -1.3231 & 0.1230 & 121.3450 & 0.4019 \\
      &TDKStain~\cite{peng2024advancing} & 11.0847 & 0.1263 & -0.4522 & 0.7451 & 72.6141 & \textbf{0.2498} & 19.6842 & 0.4652 & -0.5041 & 0.5925 & 50.3415 & 0.2529 \\
      &PSPStain~\cite{chen2024pathological} & 11.1210 & 0.1213 & -0.5439 & 0.7022 & 57.0579 & 0.2585 & 18.6233 & 0.4240 & -0.2761 & 0.7651 & 39.6011 & 0.2586 \\
           
        &DDBM~\cite{zhou2023denoising} & \textbf{11.7847} & 0.1136 & -1.2303 & -0.0095 & 206.7112 & 0.4091 & 20.5214 & 0.4749 & -1.3201 & 0.0103 & 290.9201 & 0.3711 \\
      &UNSB~\cite{kim2023unpaired} & 11.0874 & 0.1392 & -0.6440 & 0.1654 & 54.7004 & 0.3155 & 19.4668 & 0.4963 & -0.6459 & 0.3130 & 61.3214 & 0.3151 \\
       \Xhline{0.8pt}
      \rowcolor{gray!10}
     & ControlNet~\cite{zhang2023adding}  & 7.9736 & 0.1028 & +0.6942 & 0.2747 & 143.5401 & 0.3846 & 7.4402 & 0.2908 & +25.7725 & 0.0412 & 235.9315 & 0.4063 \\
      \rowcolor{gray!10}
     & VIMs~\cite{dubey2024vims} & 11.3224 & 0.1226 & -1.0667 & 0.0945 & 86.0438 & 0.3693 & 17.4837 & 0.4001 & -0.8389 & 0.3534 & 90.6424 & 0.3068 \\
      \rowcolor{gray!10}
      &  UMDST~\cite{lin2022unpaired} & 
   11.1742 & \textbf{0.1449} & -0.3930 & 0.4135 & 65.5618 & 0.3183 &
   20.5402 & 0.5346 & -1.1084 & 0.1270 & 63.3261 & 0.3448 \\



     
     \rowcolor{gray!10}
    \multirow{-4}{*}{One-to-many}&  PGVMS & 11.0508 & 0.1234 & \textbf{-0.3262} & \textbf{0.7751} & \textbf{53.3853} & 0.2552 & 19.7848 & 0.4750 & \textbf{-0.1098} & \textbf{0.7940} & \textbf{36.8306} & \textbf{0.2326} \\
      \hline
      
       \multirow{3}{*}{Type} &\multirow{3}{*}{Method} & 
      \multicolumn{6}{c|}{PR} & \multicolumn{6}{c}{Ki67} \\
      \cline{3-14}
      && \multicolumn{2}{c|}{Quality (reference)} & \multicolumn{2}{c|}{Pathology} & \multicolumn{2}{c|}{Perception} & 
      \multicolumn{2}{c|}{Quality (reference)} & \multicolumn{2}{c|}{Pathology} & \multicolumn{2}{c}{Perception} \\
      \cline{3-14}
      && PSNR$\uparrow$ & SSIM$\uparrow$ & IOD$_{\times10^7}$ & Pearson-R$\uparrow$ & FID$\downarrow$ & DISTS$\downarrow$ & PSNR$\uparrow$ & SSIM$\uparrow$ & IOD$_{\times10^7}$ & Pearson-R$\uparrow$ & FID$\downarrow$ & DISTS$\downarrow$ \\
      \hline
     \multirow{9}{*}{One-to-one} 
     & CycleGAN~\cite{zhu2017unpaired} & 17.9551 & 0.4544 & +1.3643 & 0.2339 & 66.0757 & 0.3113 & 18.7757 & 0.3840 & +0.8227 & 0.6277 & 52.5754 & 0.2328 \\
     & Pix2Pix~\cite{isola2017image} & 23.9329 & 0.5556 & -0.5986 & 0.1558 & 116.1813 & 0.2702 & 20.5521 & 0.4413 & -0.4839 & 0.2138 & 140.2222 & 0.2914 \\
      &CUT~\cite{park2020contrastive} & 19.0278 & 0.4882 & +0.6367 & 0.2133 & 47.5413 & 0.3153 & 17.2823 & 0.3017 & -0.5802 & 0.1187 & 114.3944 & 0.3063 \\
     & ASP~\cite{li2023adaptive} & 23.3897 & 0.5454 & -0.5677 & -0.0954 & 42.3880 & 0.2575 & 19.1923 & 0.3849 & -0.2544 & 0.5686 & 27.0728 & 0.2272 \\
      &PyramidP2P~\cite{liu2022bci} & 24.1510 & \textbf{0.6187} & -0.5892 & 0.0479 & 87.7714 & 0.3378 & 20.7354 & \textbf{0.4895} & -0.4475 & 0.3627 & 63.2432 & 0.3150 \\
    &  TDKStain~\cite{peng2024advancing} & 21.9699 & 0.5236 & -0.2936 & 0.0073 & 64.9066 & 0.2739 & 19.4341 & 0.4014 & \textbf{-0.0994} & 0.8735 & 38.1665 & 0.2302 \\
    &  PSPStain~\cite{chen2024pathological} & 23.8481 & 0.5544 & -0.5586 & 0.5877 & 43.8625 & 0.2503 & 18.9546 & 0.3966 & +0.4443 & 0.8744 & 32.2626 & 0.2539 \\

    &  DDBM~\cite{zhou2023denoising} & 22.5329 & 0.5293 & -5.6605 & 0.0136 & 239.2746 & 0.3284 & 19.1372 & 0.3744 & -0.3313 & 0.5537 & 120.2547 & 0.3166 \\
     & UNSB~\cite{kim2023unpaired} & 23.7180 & 0.5957 & -0.5131 & 0.2865 & 42.2145 & 0.3052 & 20.5115 & 0.4523 & -0.4590 & 0.7595 & 35.9533 & 0.2905 \\
        \Xhline{0.8pt}
      \rowcolor{gray!10}
     & ControlNet~\cite{zhang2023adding}  & 8.3349 & 0.2686 & +9.6558 & 0.0084 & 259.7857 & 0.4424 & 8.8441 & 0.2168 & +6.5673 & 0.6417 & 252.8057 & 0.4164 \\
     \rowcolor{gray!10}
    &  VIMs~\cite{dubey2024vims} & \textbf{24.2096} & 0.5438 & -0.6146 & 0.0075 & 112.9235 & 0.3357 & \textbf{20.7378} & 0.3930 & -0.5592 & 0.5227 & 84.4047 & 0.3316 \\
      \rowcolor{gray!10}
       &  UMDST~\cite{lin2022unpaired}&
    22.5452 & 0.5795 & -0.4066 & 0.0761 & 45.6025 & 0.3120 &
    20.1732 & 0.4651 & -0.4733 & 0.5237 & 56.5546 & 0.3003 \\
    
    \rowcolor{gray!10}




    \multirow{-4}{*}{One-to-many} & PGVMS & 22.4371 & 0.5388 & \textbf{-0.0499} & \textbf{0.7084} & \textbf{40.6377} & \textbf{0.2495} & 20.3849 & 0.4191 & -0.2256 & \textbf{0.8913} & \textbf{24.4450} & \textbf{0.2208} \\
      \Xhline{1.5pt} 
    \end{tabular}
  }
\end{table*}

\subsection{Implementation Details}
We implement our framework using PyTorch on an NVIDIA RTX A6000 GPU. Building upon the CUT framework~\cite{park2020contrastive}, we employ a ResNet-6Blocks architecture for the generator and a PatchGAN discriminator~\cite{isola2017image}. The model is trained with random $512 \times 512$ patches using a batch size of 1. Optimization is performed using Adam with a fixed learning rate of $1 \times 10^{-4}$ for 80 epochs. The loss weighting parameters are set as follows: $\lambda_{M} = 1.0$, $\lambda_{C} = 2.5$, $\lambda_{S} = 0.05$, and $\lambda_{G} = 10.0$.

{During training, the uniplex IHC data are used as image-text pairs to train the prompt-guided virtual staining model. we set the batch size to 1 to ensure that each single-stain IHC image is paired with its corresponding stain-type text label (e.g., ``HER2 stain'', ``Ki67 stain'', etc.) and input simultaneously into the language model. This setup allows the model to learn from multiple one-to-one image–text pairs during training, while ultimately enabling a one-to-many virtual staining capability at inference.}

\subsection{Evaluation Metrics}
Our validation employs a three-level quantitative assessment framework evaluating image quality, pathological correlation, and perceptual consistency. At the image quality level, we measure peak signal-to-noise ratio (PSNR) and structural similarity index measure (SSIM). The pathological correlation level assesses integrated optical density (IOD) \cite{zhang2022mvfstain} and pearson correlation coefficient (Pearson-R) \cite{liu2021unpaired}, while the perceptual consistency level evaluates Fréchet inception distance (FID) and deep image structure and texture similarity (DISTS).
 

\textit{1) SSIM:}
In image translation tasks, SSIM is often used to evaluate image quality, primarily to measure the similarity of different images in brightness, contrast, and structure. 



\textit{2) PSNR:}
As a widely used objective image quality metric, PSNR measures the differences between images by comparing the ratio of signal to noise.



\textit{3) IOD:} IOD quantifies the difference in total protein expression between generated ($test$) and GT ($label$) images.
\begin{equation}
IOD = \sum_{i \in \Omega_{test}} OD_i^{test} - \sum_{i \in \Omega_{label}} OD_i^{label}
\end{equation}
where $\Omega$ denotes dataset, $OD_i$ represents optical density at image $i$. 

\textit{4) Pearson-R:}
Pearson-R evaluates protein expression pattern consistency between generated image and label.
\begin{equation}
\footnotesize
Pearson\text{-}R = \frac{\sum_{i=1}^I (OD_i^{test} - \bar{OD}^{test})(OD_i^{label} - \bar{OD}^{label})}{\sqrt{\sum_{i=1}^I (OD_i^{test} - \bar{OD}^{test})^2 \sum_{i=1}^I (OD_i^{label} - \bar{OD}^{label})^2}}
\end{equation}

\textit{5) FID:} FID quantifies the similarity between generated and real images by comparing their feature from Inception V3. 


\textit{6) DISTS: }
DISTS is a deep learning-based perceptual metric that quantifies image similarity by comparing multi-level structural and textural features extracted through VGG16.





\subsection{Comparative Results and Analysis}
 In this section, we compare the performance of our method to other SOTA methods quantitatively and qualitatively based on the MIST and IHC4BC datasets. 
{The compared methods can be
 broadly classified into two parts: (1) One-to-one virtual staining methods, including Pix2Pix \cite{isola2017image}, CycleGAN \cite{zhu2017unpaired}, CUT \cite{park2020contrastive}, PyramidP2P
 \cite{liu2022bci}, ASP \cite{li2023adaptive}, TDKStain \cite{peng2024advancing} , PSPStain \cite{chen2024pathological}, DDBM \cite{zhou2023denoising} and UNSB \cite{kim2023unpaired};  (2) One-to-many virtual staining methods, including ControlNet \cite{zhang2023adding}, VIMs \cite{dubey2024vims} and UMDST \cite{lin2022unpaired}.}

 






\subsubsection{Quantitative Assessment}

\begin{figure*}[!htbp]
\centerline{\includegraphics[width=\textwidth]{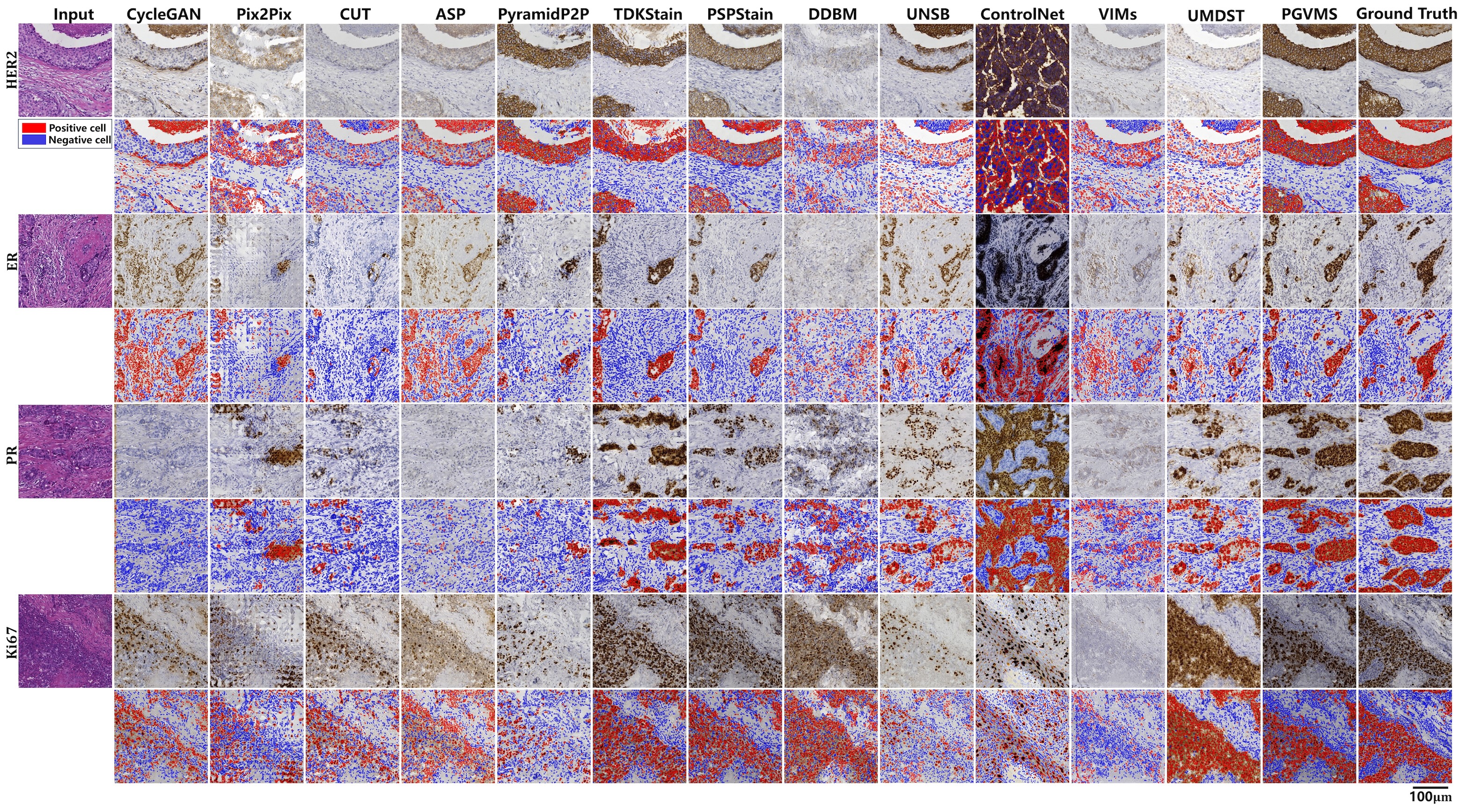}}
\caption{{Virtual IHC-stained image results of different methods on the MIST dataset. The first column is the H\&E-stained images. Columns 2 to 14 are the images virtually stained by different methods. The last column is the ground truth. The figure also demonstrates virtual IHC results alongside positive cell visualization from DeepLIIF.}}
\label{qualitative_1}
\end{figure*}

Empirical results (Tables \ref{tab:all_datasets} and \ref{tab:ihc4bc_results}) demonstrate our method's superiority in pathological correlation and perceptual consistency, 
while conventional image quality metrics (PSNR and SSIM) show limited improvement due to inherent structural mismatches between the two staining modalities in the dataset\cite{liu2022bci}.
As detailed in Section \ref{sec:discussion} with Table~\ref{tab:combined_psnrssim_results}, PSNR and SSIM provide only reference values rather than conclusive evaluation criteria.


 We then evaluate pathological correlation (IOD and Pearson-R) in terms of variation in total protein expression levels and consistency of expression changes, respectively. For total protein expression levels, as shown in Tables \ref{tab:all_datasets} and \ref{tab:ihc4bc_results} pathology columns, huge variation in IOD is obtained with Pix2Pix, ASP, ControlNet and PyramidP2P, demonstrating their weakness in preserving protein expression from the source images \cite{zhang2022mvfstain}. While other models show better preservation of protein expression levels in stain transfer, our method achieves the best IOD. Furthermore, PGVMS achieves the highest Pearson-R score in both datasets, and this indicates that each generated image closely resembles the ground truth in terms of pathological correlation. These superior IOD and Pearson-R performance metrics of PGVMS can be attributed to the innovative components: PALS maintains protein expression levels matching ground truth, preserving pathological semantics, while PSSG integrates CONCH's semantic embeddings during stain transfer to ensure structural and biochemical fidelity.
 

 Finally, we evaluate the visual perception (FID and DISTS). As shown in Tables \ref{tab:all_datasets} and \ref{tab:ihc4bc_results} perception columns, our method achieved the lowest FID and DISTS metrics, demonstrating the most realistic stain transformation and tissue structure preservation for pathological diagnosis. This performance advantage stems from PCLS's design which enforces consistent feature embedding in latent space, thereby preserving key pathological features in synthesized protein margins and cellular atypia.


\begin{figure*}[t]
\centerline{\includegraphics[width=\textwidth]{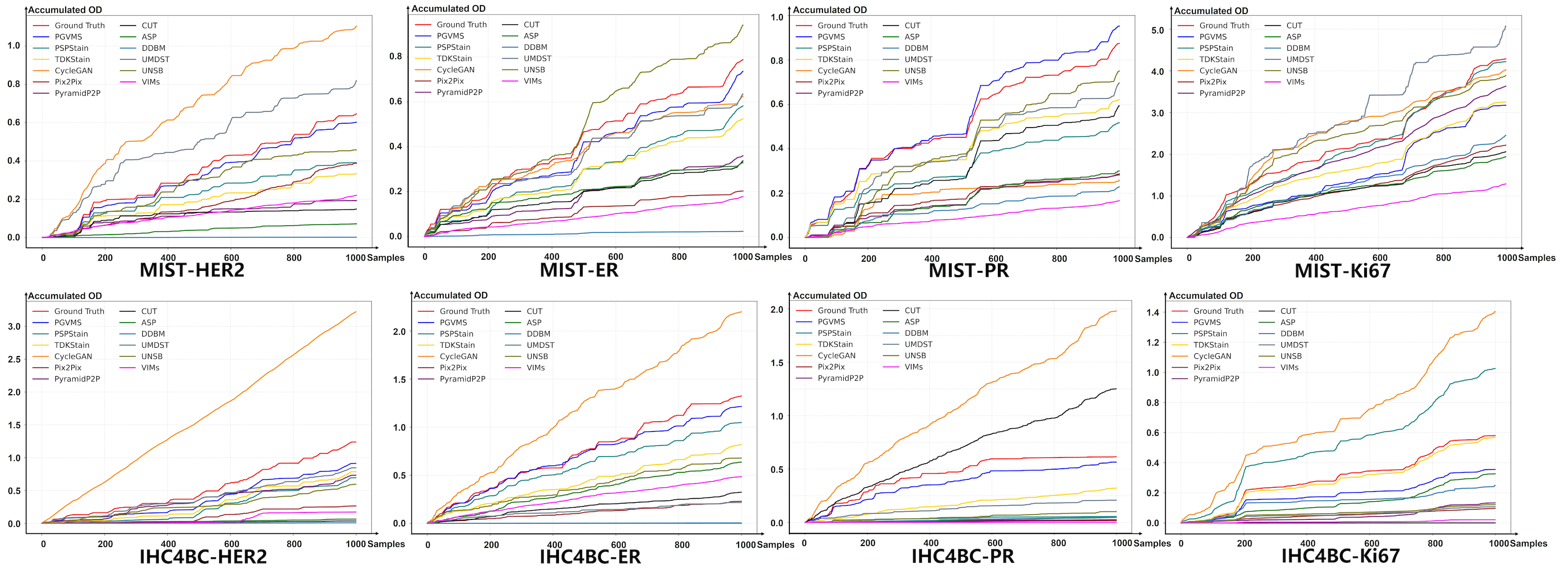}}
\caption{{Consistent with the protein progression tendency. The horizontal axis (abscissa) represents the sample index, while the vertical axis (ordinate) denotes the accumulated optical density (OD) value. The figure illustrates the accumulated OD curves of various virtual staining methods for IHC positive regions, comparing the virtual results with the reference IHC images on the MIST and IHC4BC datasets. The results of the ControlNet model are excluded because they deviate significantly from the ground truth.}}
\label{qualitative_2}
\end{figure*}

\begin{figure*}[!htbp]
\centerline{\includegraphics[width=\textwidth]{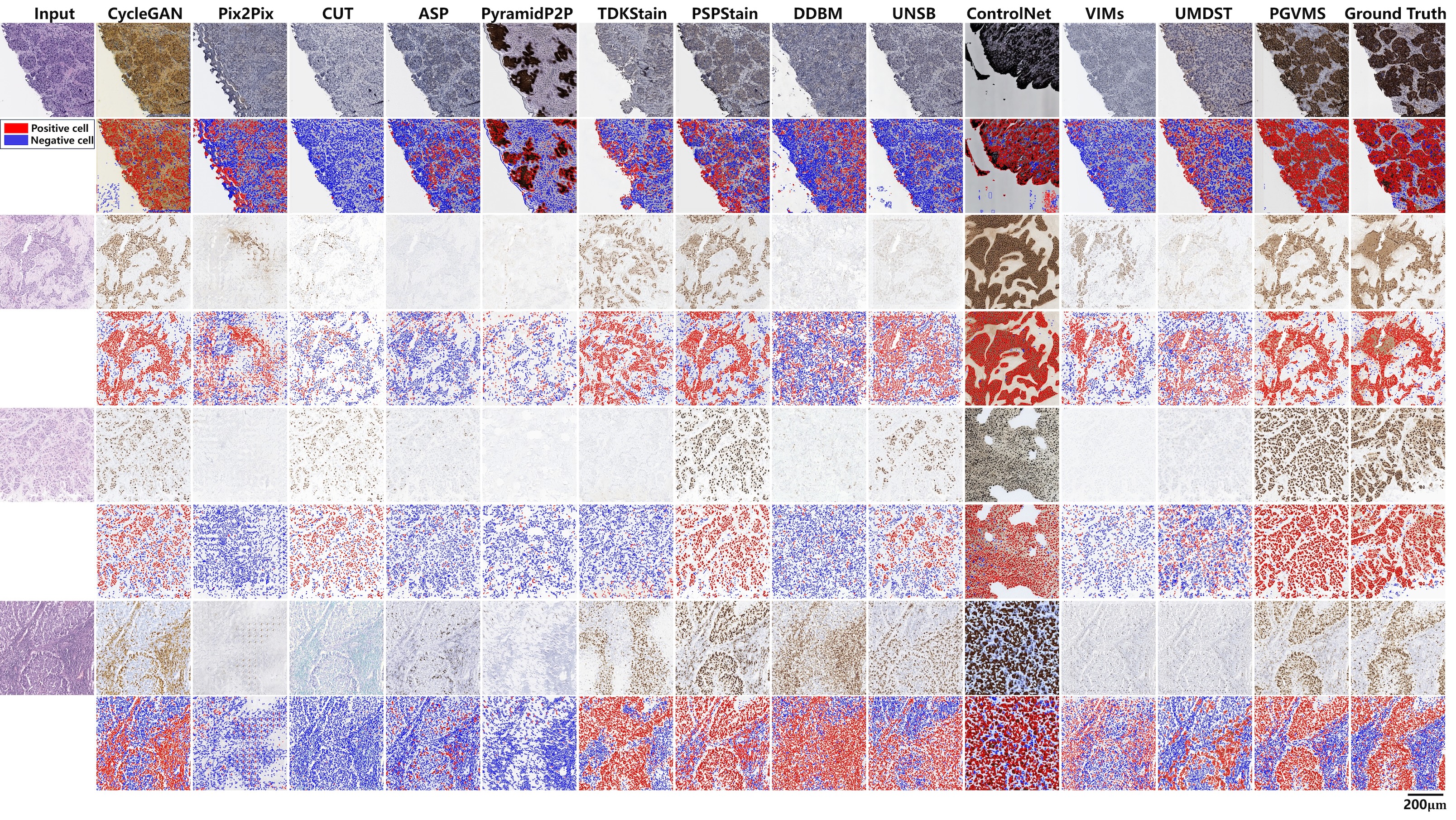}}
\caption{{Virtual IHC-stained image results of different methods on the IHC4BC dataset. The first column is the H\&E-stained images. Columns 2 to 14 are the images virtually stained by different methods. The last column is the ground truth. The figure also demonstrates virtual IHC results alongside positive cell visualization from DeepLIIF.}}
\label{qualitative_ihc4bc}
\end{figure*}

\subsubsection{Qualitative Assessment} To further explore the visual differences in the virtual staining, we present the results of our method and other competing methods on MIST dataset in Fig. \ref{qualitative_1} and IHC4BC dataset in Fig. \ref{qualitative_ihc4bc}. {DeepLIIF \cite{ghahremani2022deep} was used to highlight positive cells (red) and negative cells (blue).} Compared with others, the virtual IHC images generated by our method are most similar with the real IHC images.

Fig. \ref{qualitative_1} and Fig. \ref{qualitative_ihc4bc} reveals that existing methods like CycleGAN, Pix2Pix, and CUT produce IHC images with inaccurate pathological characteristics due to insufficient supervisory information. Moreover, ASP and PyramidP2P contain insufficient pathological characteristics due to the patch-level loss between virtual IHC images and weakly paired ground truth. {ControlNet often produces hallucinations due to the lack of sufficient pathological constraints. VIMs and UMDST, as one-to-many methods, lack the precision to identify protein overexpression regions accurately.}
In contrast, both TDKStain and PSPStain detect the protein expression amount, thus they maintain strong pathological relevance with real IHC images.  However, TDKStain often produces wrong tumor shapes and distributions (Fig. \ref{tdkstain_unpair}), as it fails to accurately capture cellular localization and may introduce spurious cellular artifacts. Meanwhile, PSPStain preserves the tissue structure in H\&E-stained images through semantic preservation. Notably, virtual IHC images generated by PGVMS exhibit closer similarity to referenced adjacent-layer IHC images by learning pathological semantics—particularly in highlighting small high-protein expression areas missed by other methods.
\subsubsection{Protein Expression Analysis}
Typically, pathologists pay more attention to the protein expression level of IHC staining when analyzing IHC images \cite{li2025biomedical}. Therefore, the accuracy of virtual IHC images can be effectively evaluated by studying the correlation between the protein expression level in virtual IHC images and those in real IHC images from adjacent layers.

For each IHC patch, the H channel and the DAB channel can be separated using deconvolution. In the DAB channel, we calculate the tumor expression level for each patch and then accumulate the expression levels across all patches \cite{chen2024pathological}, as shown in the line chart (Fig. \ref{qualitative_2}). The red line in the chart represents the cumulative curve of positive protein expression levels for the reference. Notably, CycleGAN (orange) exhibits substantial deviation from the reference (red) trajectory, indicating poor pathological concordance. {The results of the ControlNet model are excluded because they deviate significantly from the ground truth.} PSPStain (cyan) and TDKStain (yellow) exhibit good consistency. However, our PGVMS (blue) demonstrates the closest approximation to the reference profile across three stain types, visually confirming its superior pathological consistency with authentic IHC results.

{
\subsubsection{Downstream Classification Experiments}
To evaluate the diagnostic value of virtual staining, we assessed the impact of generated IHC images on downstream biomarker classification (Table~\ref{tab:combined_classification_full_2}) using a ViT-B/32 backbone classifier. Separate models were trained for each biomarker—HER2, ER, PR, and Ki67—on the MIST and IHC4BC datasets to measure how well pathological information was preserved.
}
\begin{table}[ht]
\centering
\caption{{Classification accuracy (\%) of virtual staining results on MIST and IHC4BC datasets. HER2: 4-class (0, 1+, 2+, 3+); ER, PR, Ki67: binary (negative/positive). Models are grouped into one-to-one and one-to-many categories.}}
\resizebox{\columnwidth}{!}{
\begin{tabular}{l|cccc|cccc}
\Xhline{1.3pt}
\multirow{2}{*}{Method} & \multicolumn{4}{c|}{MIST} & \multicolumn{4}{c}{IHC4BC} \\
\cline{2-9}
 & HER2 & ER & PR & Ki67 & HER2 & ER & PR & Ki67 \\
\hline
CycleGAN    & 0.453     & 0.570     & 0.534    & 0.767     & 0.495     & 0.600     & 0.479     & 0.771     \\
Pix2Pix     & 0.490 & 0.671 & 0.671 & \underline{0.847} & 0.422 & 0.595 & 0.666 & 0.613 \\
CUT         & 0.457     & 0.610     & 0.664     & 0.807     & 0.258     & 0.573     & 0.492     & 0.565     \\
ASP         & 0.378 & 0.623 & 0.637 & 0.759 & 0.256 & 0.547 & 0.559 & 0.799 \\
PyramidP2P  & 0.587 & 0.695 & 0.705 & 0.842 & 0.505 & 0.497 & 0.640 & 0.686 \\
TDKStain    & \underline{0.642} & \textbf{0.798} & \underline{0.812} & \textbf{0.857} & \textbf{0.541} & \textbf{0.667} & 0.571 & \underline{0.834} \\
PSPStain    & \textbf{0.647} & 0.763 & 0.768 & 0.844 & 0.474 & 0.596 & \textbf{0.750} & 0.763 \\
DDBM        & 0.420 & 0.584 & 0.577 & 0.634 & 0.449 & 0.534 & 0.629 & 0.754 \\
UNSB        & 0.450 & 0.642 & 0.611 & 0.802 & 0.397 & 0.563 & 0.626 & 0.782 \\
\Xhline{0.8pt}
\rowcolor{gray!10}
ControlNet  & 0.329 & 0.673 & 0.537 & 0.844 & 0.470 & 0.558 & 0.352 & 0.599 \\
\rowcolor{gray!10}
VIMs        & 0.309 & 0.652 & 0.561 & 0.795 & 0.326 & 0.626 & 0.657 & 0.578 \\
\rowcolor{gray!10}
UMDST       & 0.419 & 0.572 & 0.655 & 0.843 & 0.251 & 0.462 & 0.640 & 0.687 \\
\rowcolor{gray!10}
PGVMS       & 0.617 & \underline{0.772} & \textbf{0.818} & 0.740 & \underline{0.535} & \underline{0.627} & \underline{0.678} & \textbf{0.865} \\
\Xhline{1.3pt}
\end{tabular}
}
\label{tab:combined_classification_full_2}
\end{table}

{
HER2 was treated as a four-class problem (0, 1+, 2+, 3+) based on cumulative OD thresholds of 500, 2000, and 5000. ER and PR were binarized at 1000, and Ki67 at 2000, following established cutoffs \cite{varghese2014ihc}. All automated labels were subsequently reviewed by board-certified pathologists.
}

\begin{figure}[!htbp]
\centerline{\includegraphics[width=\columnwidth]{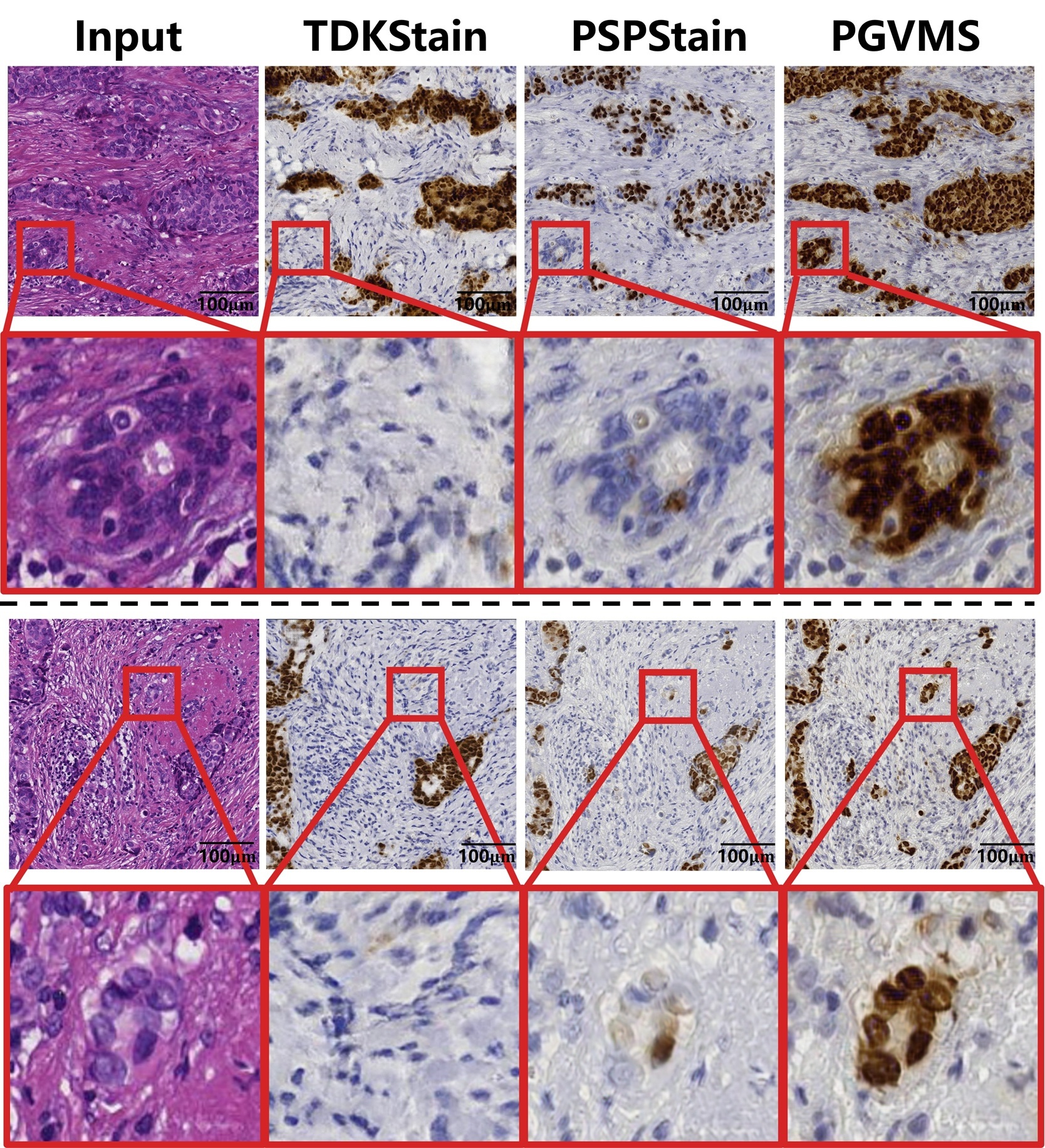}}
\caption{{Visual comparison illustrating the severe cell spatial misalignment issue in TDKStain results, where virtual staining causes substantial cell displacement relative to the input H\&E image. In contrast, our PGVMS method preserves accurate cell spatial localization.}}
\label{tdkstain_unpair}
\end{figure}

{
As shown in Table~\ref{tab:combined_classification_full_2}, TDKStain achieves the highest classification accuracy, followed by PGVMS and PSPStain. However, TDKStain shows severe cell localization errors, with notable misalignment between the generated IHC and input H\&E images (Figure~\ref{tdkstain_unpair}), significantly limiting its clinical usability.
}

\subsubsection{Feature-Level Metric Based on Pathological Foundation Model CONCH}
{
To assess the feature-level alignment of virtual IHC images with clinical pathology, we encode both the generated and ground-truth IHC images using the pre-trained image encoder of CONCH, producing a 512-dimensional feature vector for each image. The CONCH-FID is then calculated as the Fréchet distance between the distributions of these embeddings, capturing how well the generated images preserve high-level, pathology-specific semantic features. As shown in Table~\ref{tab:combined_fid_conch_5}, our PGVMS method ranks second on the MIST dataset, first on the IHC4BC dataset, and achieves the best overall performance across datasets. In comparison, PSPStain ranks second overall, while ASP comes third. These results highlight PGVMS’s superior ability to generate virtual IHC images that closely replicate the structural and semantic characteristics of authentic clinical slides.
}
\begin{table*}[!htbp]
\centering
\caption{{Feature-level FID between virtual IHC and ground-truth IHC computed using the pathology foundation model CONCH. Lower values indicate better alignment. The best and second-best results are highlighted in \textbf{bold} and \underline{underlined}, respectively.}}
\label{tab:combined_fid_conch_5}
\setlength{\tabcolsep}{4pt}
\renewcommand{\arraystretch}{1.15}
\resizebox{\textwidth}{!}{
\begin{tabular}{l|l|cccc|c||cccc|c}
\Xhline{1.3pt}
Type & Method 
& \multicolumn{5}{c||}{\textbf{MIST}} 
& \multicolumn{5}{c}{\textbf{IHC4BC}} \\
\cline{3-12}
& & HER2 & ER & PR & Ki67 & Mean 
& HER2 & ER & PR & Ki67 & Mean \\
\hline
\multirow{9}{*}{One-to-one}
& CycleGAN   & 102.1581 & 52.9350 & 125.7399 & 48.0381 & 82.2178 & 218.4938 & 44.1311 & 81.4923 & 49.0361 & 98.2884 \\
& Pix2Pix    & 95.9983 & 189.5522 & 182.9658 & 216.5276 & 171.2610 & 153.2364 & 105.1925 & 95.7887 & 128.3961 & 120.6534 \\
& CUT        & 62.6960 & 39.4071 & \underline{32.4882} & 35.4223 & 42.5034 & 80.8654 & 32.3410 & 39.4285 & 247.6296 & 100.0661 \\
& ASP        & \underline{39.9833} & 52.4421 & 38.3425 & 43.7905 & 43.6396 & 76.8791 & \underline{25.4769} & \underline{31.9873} & \textbf{20.9709} & \underline{38.8285} \\
& Pyramid    & 151.2503 & 153.7980 & 145.9719 & 172.3845 & 155.8512 & 237.9637 & 172.1971 & 162.0819 & 106.7118 & 169.7386 \\
& TDKStain   & 85.3784 & 73.2279 & 98.2533 & 143.9924 & 100.2130 & \underline{49.5412} & 61.5834 & 75.3190 & 52.3977 & 59.7103 \\
& PSPStain   & \textbf{34.7499} & \textbf{30.8958} & 32.7645 & \textbf{26.2759} & \textbf{31.1715} & \textbf{47.1362} & 35.2759 & 48.1594 & 36.9802 & 41.8879 \\
& DDBM       & 291.1494 & 281.9772 & 244.6667 & 215.1843 & 258.2444 & 326.7613 & 256.5512 & 258.2804 & 182.5962 & 256.0473 \\
& UNSB       & 61.9792 & 38.1701 & 49.4495 & 36.9417 & 46.6351 & 64.6003 & 69.6588 & 37.2139 & 48.7528 & 55.0565 \\
\hline
\rowcolor{gray!10}

& ControlNet & 283.2979 & 253.5201 & 317.4814 & 490.4310 & 336.1826 & 322.7134 & 432.5187 & 455.0586 & 434.0198 & 411.0776 \\
\rowcolor{gray!10}
& VIMs       & 353.5241 & 240.2728 & 234.2091 & 233.9704 & 265.4941 & 192.7621 & 140.0983 & 177.3091 & 166.0101 & 169.0449 \\
\rowcolor{gray!10}
& UMDST      & 154.3398 & 168.4891 & 147.7141 & 254.2693 & 181.2031 & 98.5943 & 116.0732 & 100.4608 & 184.7146 & 124.9607 \\

\rowcolor{gray!10}
\multirow{-4}{*}{One-to-many}& PGVMS 
& 55.1115 & \underline{31.0460} & \textbf{23.5529} & \underline{29.5545} & \underline{34.8162} 
& 61.7747 & \textbf{24.5485} & \textbf{30.8413} & \underline{21.8956} & \textbf{34.7650} \\
\Xhline{1.3pt}
\end{tabular}
}
\end{table*}

\subsubsection{Pathologist-Based Image Quality Evaluation}
\begin{figure}[!htbp]
\centerline{\includegraphics[width=\columnwidth]{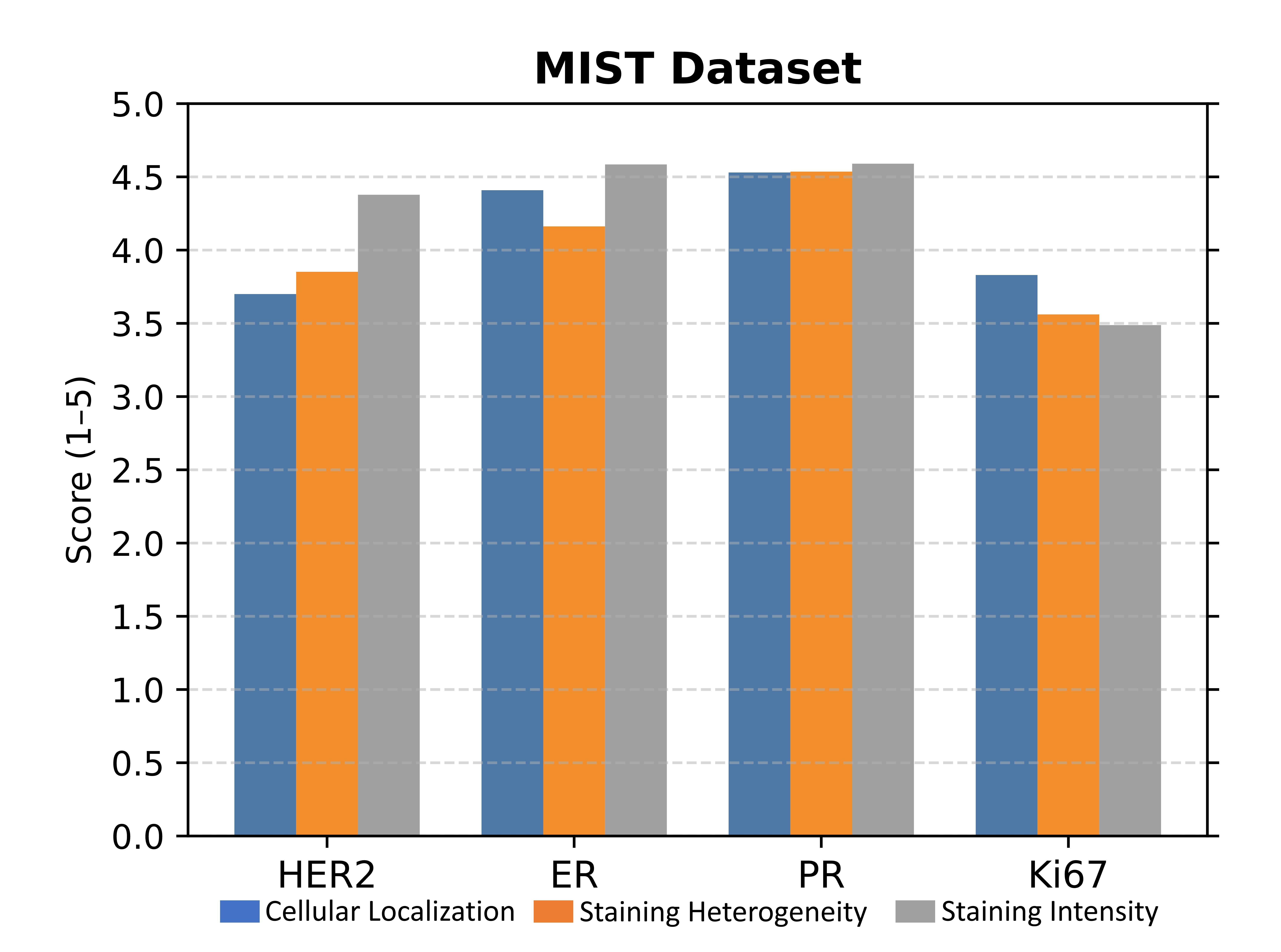}}
\caption{{Pathologist evaluation of virtual IHC-stained images on the MIST datasets across three criteria: cellular localization, staining heterogeneity, and staining intensity.}}
\label{qualitative_path_MIST}
\end{figure}

\begin{figure}[!htbp]
\centerline{\includegraphics[width=\columnwidth]{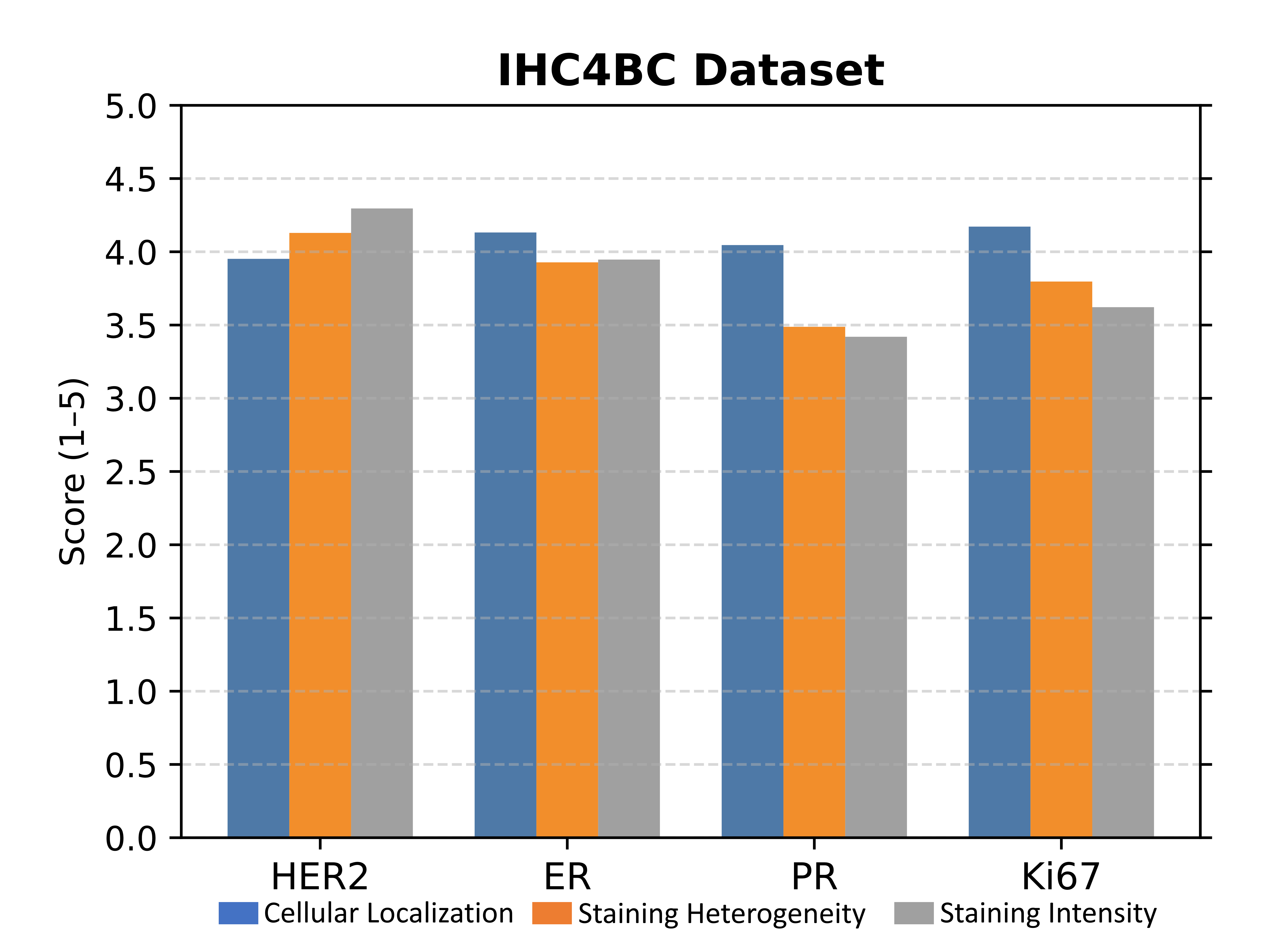}}
\caption{{Pathologist evaluation of virtual IHC-stained images on the IHC4BC datasets across three criteria: cellular localization, staining heterogeneity, and staining intensity.}}
\label{qualitative_path_IHC4BC}
\end{figure}

{
To assess the clinical feasibility of our virtual staining results, we invited experienced pathologists to conduct a subjective quality evaluation. Specifically, for each dataset (MIST and IHC4BC), 50 randomly selected patches per stain were scored on a 1–5 scale based on three criteria: cellular localization, staining heterogeneity, and staining intensity. Scores of 3 and above indicate acceptable quality, while scores below 3 are considered unsatisfactory.
}
{
As illustrated in Fig.~\ref{qualitative_path_MIST} and Fig.~\ref{qualitative_path_IHC4BC}, the virtual IHC-stained images achieve consistently high quality across all criteria, with mean scores exceeding 3.0 and often approaching 4.0 or higher. On the MIST dataset, ER and PR stains receive the highest scores in both localization and intensity, suggesting that the model effectively preserves nuclear and cytoplasmic structures. On the IHC4BC dataset, HER2 and Ki67 exhibit stable performance across all aspects, demonstrating good generalization to cross-domain and inter-patient variations.
}
{
Overall, these results confirm that our method generates visually convincing and histologically consistent virtual IHC images, which are well accepted by pathologists and hold strong potential for clinical application.
}

\subsubsection{Unified Multiple IHC Virtual Staining Analysis}
\begin{figure}[t]
\centerline{\includegraphics[width=\columnwidth]{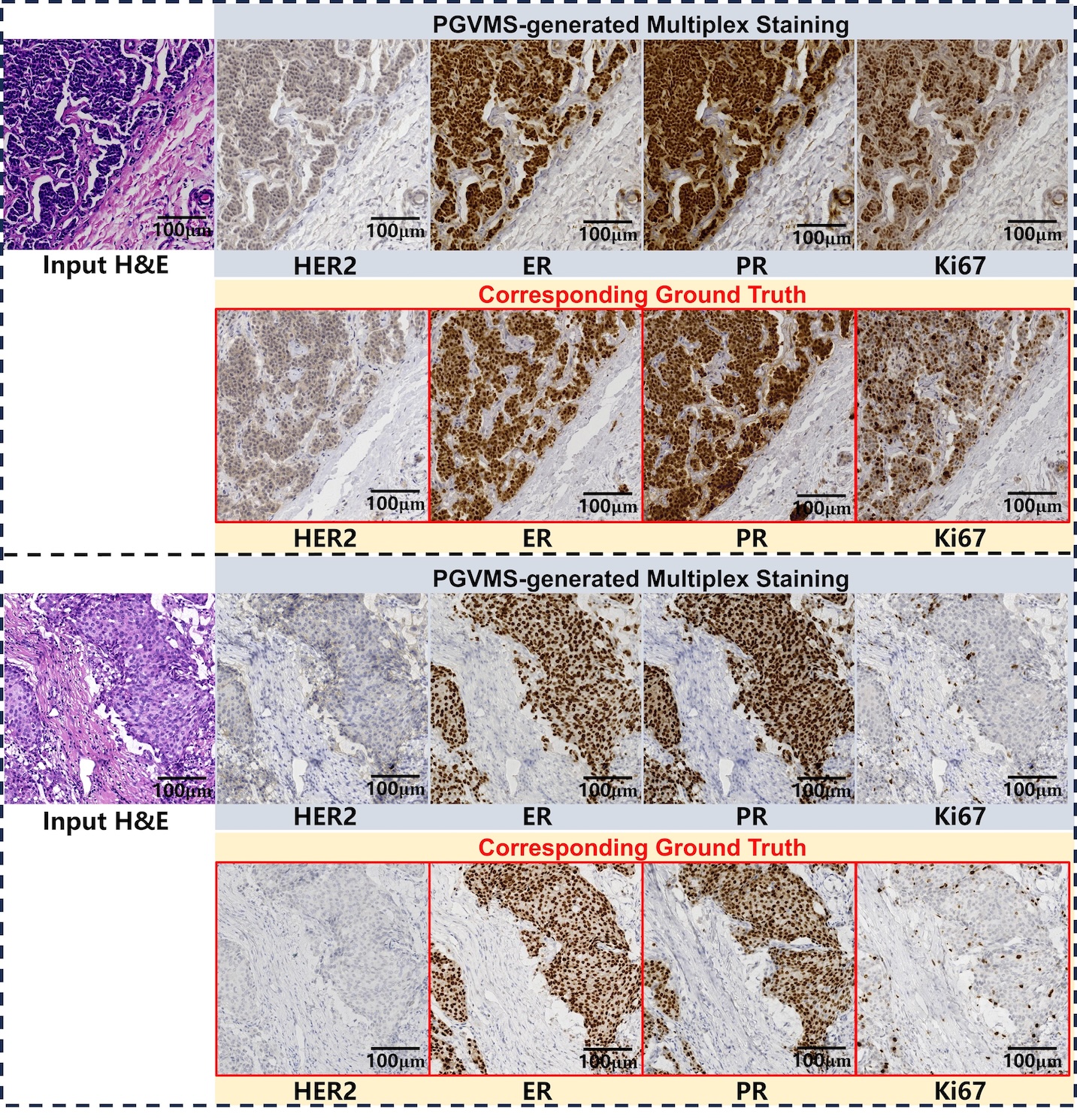}}
\caption{\revision{Virtual multiplex IHC generation from H\&E-stained input images. For each H\&E input, the first row (gray background) displays PGVMS-generated staining for HER2, ER, PR, and Ki-67, and the second row (yellow background) shows the corresponding ground-truth IHC images.}}

\label{qualitative_3}
\end{figure}

{Fig.~\ref{qualitative_3} shows the virtual generation of HER2, ER, PR, and Ki67 stained images from H\&E images, with each stain exhibiting distinct characteristics—for example, the first image shows strong ER, PR, and Ki67 but minimal HER2 expression, while the second shows strong ER and PR with low HER2 and Ki67. These results demonstrate that our method can generate multiple IHC stains simultaneously, aiding breast cancer subtype identification. The multiplex IHC images used for visualization are from the MIST dataset, which is mostly uniplex (\(\sim 77.8\%\)), with only a small fraction containing 2–4 stains.
}

\subsection{Ablation Study and Analysis}

\subsubsection{Module Ablation}
The proposed PGVMS accurately transforms H\&E-stained images into IHC-stained images through three core mechanisms: PALS, PCLS, and the PSSG generator. \revision{The baseline, built on the CUT framework, serves as an initial model for stain translation.} PALS ensures pathological consistency by preserving protein activation patterns, while PCLS enhances protein realism by enforcing semantic coherence across images. The PSSG generator enables prompt-driven adaptation for multi-stain tasks. Ablation experiments on the MIST dataset (Table \ref{tab:ablation_2}, pathology columns) reveal that PALS exerts a greater impact on pathological consistency. However, with PALS alone, protein characterization remains suboptimal. Consequently, PCLS improves perception metrics and enhances protein realism (Table \ref{tab:ablation_2}, perception columns) . When combined with PSSG’s semantic-embedded prompts, these components collectively achieve optimal stain transfer results, demonstrating PGVMS’s state-of-the-art virtual staining capability.

\begin{figure}[t]
\centerline{\includegraphics[width=\columnwidth]{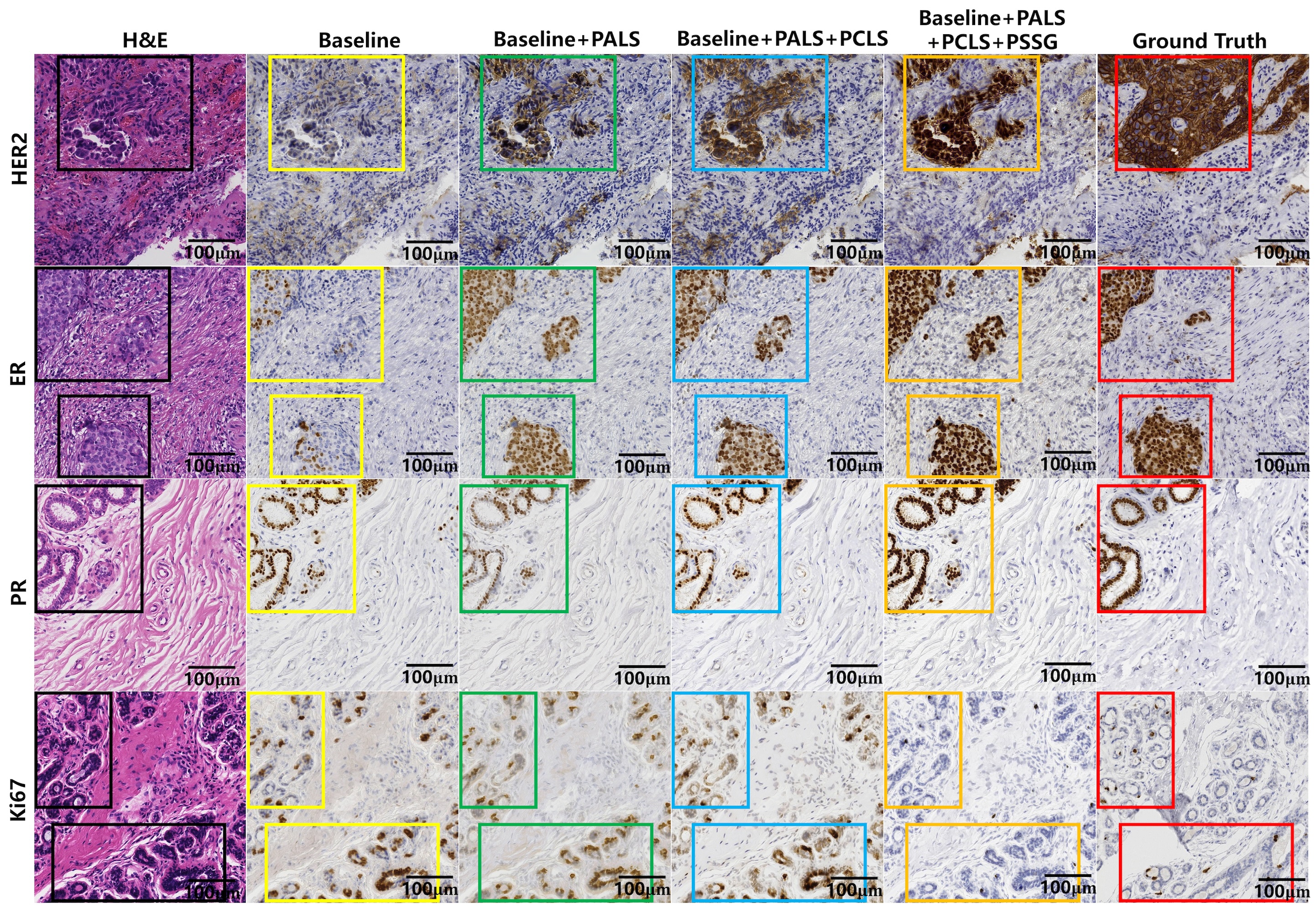}}
\caption{\revision{Visualization of the effects of PALS, PCLS, and PSSG on our PGVMS. 
The color-coded rectangles indicate regions corresponding to different experimental configurations: black for the input H\&E image, yellow for the baseline output, green for baseline + PALS, blue for baseline + PALS + PCLS, orange for baseline + PALS + PCLS + PSSG, and red for the ground truth.}}
\label{ablation_1}
\end{figure}


      
      


\begin{table}[!htbp]
  \centering
  \setlength\tabcolsep{2pt}
  \renewcommand{\arraystretch}{1.0}
  \caption{{ABLATION STUDY OF STAIN TRANSFER PERFORMANCE ON MIST DATASET. THE BEST SCORES ARE HIGHLIGHTED IN BOLD.}}
  \label{tab:ablation_2}

  \resizebox{\columnwidth}{!}{
    \begin{tabular}{cccc|cc|cc|cc}
      \Xhline{1.5pt} 
      \multirow{2}{*}{Baseline} & \multirow{2}{*}{PALS} & \multirow{2}{*}{PCLS} & \multirow{2}{*}{PSSG} 
      & \multicolumn{2}{c|}{Quality} & \multicolumn{2}{c|}{Pathology} & \multicolumn{2}{c}{Perception} \\
      \cline{5-10}
      & & & & PSNR$\uparrow$ & SSIM$\uparrow$ & IOD$_{\times10^7}$ & Pearson-R$\uparrow$ & FID$\downarrow$ & DISTS$\downarrow$ \\
      \hline
       $\checkmark$  &  &  & &
      14.2183 & 0.2046 & -3.1237 & 0.6229 & 45.6044 & 0.2598 \\
      $\checkmark$  & $\checkmark$ &  & &
      \textbf{14.6252} & \textbf{0.2205} & -2.3128 & 0.8436 & 49.7711 & 0.2577 \\
      
     $\checkmark$  & $\checkmark$ & $\checkmark$ & &
      14.4773 & 0.2195 & -2.0626 & 0.8361 & 39.2908 & 0.2508 \\
      
      $\checkmark$  &  &  & $\checkmark$&
         13.7738  &0.1905   &-5.0098   & 0.3756 & 62.5494&0.2709 \\
       $\checkmark$  & $\checkmark$ &  & $\checkmark$&
      13.0808 & 0.1887  & -2.9507 &0.3998  &67.4985  &0.2813  \\

      $\checkmark$  & $\checkmark$ & $\checkmark$ & $\checkmark$&
      14.3421 & 0.2032 & \textbf{-0.3223} & \textbf{0.8649} & \textbf{38.0654} & \textbf{0.2343} \\
      \Xhline{1.5pt} 
    \end{tabular}
  }
\end{table}
To further explore the visual effects of the PCLS, PALS, and PSSG generator  on virtual IHC image generation, we present qualitative ablation results in Fig. \ref{ablation_1}. As shown in Fig. \ref{ablation_1}, without the constraints of PALS and PCLS, the model fails to fully utilize the supervisory information provided by adjacent layer IHC images (yellow), resulting in inaccurate pathological characteristics, such as misstained nuclei or cell membranes. PALS effectively maintains the pathological consistency of the generated images by preserving molecular-level protein activation patterns (green), ensuring that the virtual staining aligns with biological reality. PCLS enhances the realism of the generated tumor regions by enforcing cross-image semantic topology consistency (blue), mitigating spatial misalignment and improving inter-image coherence. PSSG generator enables prompt-driven semantic embedding adaptation across multiple staining tasks (orange), enhancing pathological semantic embedding. When PALS and PCLS are added, the performance improves, but some staining inaccuracies remain, and certain nuclei are still incorrectly stained. When the PSSG generator, PALS, and PCLS are applied simultaneously, the virtual IHC images closely resemble the referenced adjacent layer IHC images (red), achieving the best staining transfer performance.



\subsubsection{Guidance Mechanism}
To evaluate how pathological semantic prompts operate within the PSSG generator, we replaced the CONCH model with one-hot encoding guidance and CLIP model guidance. As shown in the Table \ref{tab:ablation_CONCH_}, the minimal IOD indicates superior protein expression consistency, while the relatively low yet high absolute Pearson-R score reflects consistent tumor expression patterns. The outstanding perception metrics demonstrate optimal visual diagnostic fidelity, highlighting CONCH's capability to deliver robust pathological semantic embeddings for virtual multiplex IHC staining within a unified model.

\begin{table}[!htbp]
  \centering
  \setlength\tabcolsep{2pt}
  \renewcommand{\arraystretch}{1.0}
  \caption{\revision{ABLATION STUDY OF GUIDANCE MECHANISM ON MIST DATASET. THE BEST SCORES ARE HIGHLIGHTED IN BOLD.}}
  \label{tab:ablation_CONCH_}

  \resizebox{\columnwidth}{!}{
    \begin{tabular}{c|cc|cc|cc}
      \Xhline{1.5pt} 
      \multirow{2}{*}{ Guidance} 
      & \multicolumn{2}{c|}{Quality} & \multicolumn{2}{c|}{Pathology} & \multicolumn{2}{c}{Perception} \\
      \cline{2-7}
      & PSNR$\uparrow$ & SSIM$\uparrow$ & IOD$_{\times10^7}$ & Pearson-R$\uparrow$ & FID$\downarrow$ & DISTS$\downarrow$ \\
      \hline
     One Hot & \textbf{14.9029} & \textbf{0.2310} & -2.1429 & \textbf{0.8729} & 38.5514 & 0.2379 \\
     CLIP & 14.2745 & 0.1895 & -1.4198 & 0.8678 & 39.2263 & 0.2454 \\
     CONCH & 14.3421 & 0.2032 & \textbf{-0.3223} & 0.8649 & \textbf{38.0654} & \textbf{0.2344} \\
      \Xhline{1.5pt} 
    \end{tabular}
  }
\end{table}

\subsubsection{Image-Guided Bias Effectiveness} {We evaluated the effect of incorporating instance-specific image-derived bias into the text prompt. As shown in Table~\ref{tab:ablation_bias_mechanism}, adding the bias consistently improves pathology, quality, and perception metrics across all biomarkers, indicating that it enhances semantic alignment and preserves domain-specific features in the virtual stains.}

\begin{table}[!htbp]
  \centering
  \setlength\tabcolsep{2pt}
  \renewcommand{\arraystretch}{1.0}
  \caption{{ABLATION STUDY OF IMAGE-GUIDED BIAS MECHANISM ON MIST DATASET. THE BEST SCORES ARE HIGHLIGHTED IN BOLD.}}
  \label{tab:ablation_bias_mechanism}

  \resizebox{\columnwidth}{!}{
    \begin{tabular}{c|cc|cc|cc}
      \Xhline{1.5pt} 
      \multirow{2}{*}{Method} 
      & \multicolumn{2}{c|}{Quality} & \multicolumn{2}{c|}{Pathology} & \multicolumn{2}{c}{Perception} \\
      \cline{2-7}
      & PSNR$\uparrow$ & SSIM$\uparrow$ & IOD$_{\times10^7}$ & Pearson-R$\uparrow$ & FID$\downarrow$ & DISTS$\downarrow$ \\
      \hline
      w/o bias & 14.0289 & \textbf{0.2099} & -2.2577 & 0.7400 & 58.2488 & 0.2482 \\
      w/ bias & \textbf{14.3421} & 0.2062 & \textbf{-0.3223} & \textbf{0.8648} & \textbf{38.0654} & \textbf{0.2344} \\
      \Xhline{1.5pt} 
    \end{tabular}
  }
\end{table}

\subsubsection{Pooling Strategy in Bias Computation}
 {We compared max pooling, average pooling, and a combination of both for extracting the image-guided bias. Table~\ref{tab:ablation_pooling_avg} shows that combining max and average pooling outperforms either operation alone, suggesting that fusing localized discriminative cues with global contextual information provides a more informative and balanced bias for guiding the virtual staining.}

\begin{table}[ht]
  \centering
  \setlength\tabcolsep{2pt}
  \renewcommand{\arraystretch}{1.0}
  \caption{{ABLATION STUDY OF POOLING STRATEGY ON MIST DATASET. THE BEST SCORES ARE HIGHLIGHTED IN BOLD.}}
  \label{tab:ablation_pooling_avg}

  \resizebox{\columnwidth}{!}{
    \begin{tabular}{c|cc|cc|cc}
      \Xhline{1.5pt} 
      \multirow{2}{*}{Pooling Strategy} 
      & \multicolumn{2}{c|}{Quality} & \multicolumn{2}{c|}{Pathology} & \multicolumn{2}{c}{Perception} \\
      \cline{2-7}
      & PSNR$\uparrow$ & SSIM$\uparrow$ & IOD$_{\times10^7}$ & Pearson-R$\uparrow$ & FID$\downarrow$ & DISTS$\downarrow$ \\
      \hline
      Only Maxpooling & 14.5682 & \textbf{0.2184} & -1.7611 & 0.7890 & 53.5903 & 0.2522 \\
      Only Avgpooling & \textbf{14.6032} & 0.2086 & -2.4378 & 0.7607 & 83.3022 & 0.2646 \\
      Max \& Avg Pooling & 14.3421 & 0.2032 & \textbf{-0.3223} & \textbf{0.8648} & \textbf{38.0654} & \textbf{0.2344} \\
      \Xhline{1.5pt} 
    \end{tabular}
  }
\end{table}
\subsubsection{Fusion Strategy in PGSN}
{We ablated the prompt guidance style normalization (PGSN) against simple addition and cross-attention between the prompt and image feature. Table~\ref{tab:ablation_fusion_1} demonstrates that PGSN substantially outperforms these alternatives by adaptively modulating text embeddings with image-derived statistics, leading to stronger semantic alignment and better preservation of pathology-specific features.}

\begin{table}[!htbp]
  \centering
  \setlength\tabcolsep{2pt}
  \renewcommand{\arraystretch}{1.0}
  \caption{\revision{ABLATION STUDY OF FUSION MECHANISM ON MIST DATASET. THE BEST SCORES ARE HIGHLIGHTED IN BOLD.}}
  \label{tab:ablation_fusion_1}

  \resizebox{\columnwidth}{!}{
    \begin{tabular}{c|cc|cc|cc}
      \Xhline{1.5pt} 
      \multirow{2}{*}{ Guidance} 
      & \multicolumn{2}{c|}{Quality} & \multicolumn{2}{c|}{Pathology} & \multicolumn{2}{c}{Perception} \\
      \cline{2-7}
      & PSNR$\uparrow$ & SSIM$\uparrow$ & IOD$_{\times10^7}$ & Pearson-R$\uparrow$ & FID$\downarrow$ & DISTS$\downarrow$ \\
      \hline
     Addition & 13.8117 & 0.1904 & -1.5385 & 0.7942 & 53.1891 & 0.2451 \\
     Cross Attention & 13.8461 & 0.2028 & -2.4021 & 0.8041 & 55.5236 & 0.2438 \\
     PGSN & \textbf{14.3421} & \textbf{0.2032} & \textbf{-0.3223} & \textbf{0.8649} & \textbf{38.0654} & \textbf{0.2344} \\
      \Xhline{1.5pt} 
    \end{tabular}
  }
\end{table}



\subsubsection{Normalization Strategy}
 {We assessed the role of instance normalization (IN) and layer normalization (LN) within PGSN. Table~\ref{tab:ablation_normalization_avg} shows that using both IN and LN achieves the best overall performance, as IN suppresses irrelevant style variations while LN stabilizes global feature distributions, resulting in more accurate and consistent virtual staining.}

\begin{table}[!htbp]
  \centering
  \setlength\tabcolsep{2pt}
  \renewcommand{\arraystretch}{1.0}
  \caption{{ABLATION STUDY OF NORMALIZATION STRATEGY ON MIST DATASET. THE BEST SCORES ARE HIGHLIGHTED IN BOLD.}}
  \label{tab:ablation_normalization_avg}

  \resizebox{\columnwidth}{!}{
    \begin{tabular}{c|cc|cc|cc}
      \Xhline{1.5pt} 
      \multirow{2}{*}{Normalization Strategy} 
      & \multicolumn{2}{c|}{Quality} & \multicolumn{2}{c|}{Pathology} & \multicolumn{2}{c}{Perception} \\
      \cline{2-7}
      & PSNR$\uparrow$ & SSIM$\uparrow$ & IOD$_{\times10^7}$ & Pearson-R$\uparrow$ & FID$\downarrow$ & DISTS$\downarrow$ \\
      \hline
      Only IN & 14.4614 & 0.2184 & -3.2292 & 0.7939 & 69.9662 & 0.2600 \\
      Only LN & \textbf{15.6425} & \textbf{0.2887} & -3.9779 & 0.7672 & 153.7774 & 0.3414 \\
      Both IN\&LN & 14.3421 & 0.2032 & \textbf{-0.3223} & \textbf{0.8649} & \textbf{38.0654} & \textbf{0.2344} \\
      \Xhline{1.5pt} 
    \end{tabular}
  }
\end{table}

\subsubsection{Sensitivity Analysis of $\alpha$ in MLPA}
To investigate the influence of $\alpha$ in MLPA, we conducted an ablation study with different values ($\alpha$=1.0, 1.4, 1.8, 2.2) in Table \ref{tab:ablation_OD}. The results demonstrate a bell-shaped performance curve, where $\alpha$=1.8 achieves optimal balance between molecular expression accuracy and histological structure preservation.

\begin{table}[ht]
  \centering
  \setlength\tabcolsep{2pt}
  \renewcommand{\arraystretch}{1.0}
  \caption{\revision{ABLATION STUDY OF FOD $\alpha$ ON MIST DATASET. THE BEST SCORES ARE HIGHLIGHTED IN BOLD.}}
  \label{tab:ablation_OD}

  \resizebox{\columnwidth}{!}{
    \begin{tabular}{c|cc|cc|cc}
      \Xhline{1.5pt} 
      \multirow{2}{*}{FOD $\alpha$} 
      & \multicolumn{2}{c|}{Quality} & \multicolumn{2}{c|}{Pathology} & \multicolumn{2}{c}{Perception} \\
      \cline{2-7}
      & PSNR$\uparrow$ & SSIM$\uparrow$ & IOD$_{\times10^7}$ & Pearson-R$\uparrow$ & FID$\downarrow$ & DISTS$\downarrow$ \\
      \hline
     1.0 & 14.6189 & 0.2115 & -2.2031 & 0.8331 & 43.8208 & 0.2395 \\
     1.4 & \textbf{14.6242} & 0.2078 & -1.7750 & 0.8598 & 38.3692 & 0.2372 \\
     1.8 & 14.3421 & 0.2032 & \textbf{-0.3223} & \textbf{0.8649} & \textbf{38.0654} & \textbf{0.2344} \\
     2.2 & 14.2708 & \textbf{0.2197} & -1.8733 & 0.8507 & 43.6559 & 0.2449 \\
      \Xhline{1.5pt} 
    \end{tabular}
  }
\end{table}

\section{Discussion}

\subsection{Pathology Foundation Models Influence}
{We evaluated several vision–language foundation models — PLIP, MUSK, and CONCH — by swapping the foundation model used for text-guided virtual staining. Both PLIP and MUSK yield reasonable results under our framework, demonstrating that the proposed text-guided virtual staining pipeline is generally stable across different foundation models. However, since PLIP and MUSK were not trained on large-scale immunohistochemistry (IHC) data, they fail to capture the domain-specific embeddings required for subtle biomarker distinctions among HER2, ER, PR, and Ki67. We excluded models such as UNI and Virchow because they provide only image encoders and lack text encoders, which are essential for our prompt-guided framework. In contrast, CONCH was pre-trained on large-scale IHC pathological images and thus encodes rich pathology-specific features that better support text-guided control over fine-grained tissue and biomarker structures. Consequently, when used as the foundation for text guidance, CONCH achieves the best feature-level consistency metrics (Table~\ref{tab:ablation_foundation_1}), highlighting the benefit of using a domain-specialized foundation model for reliable text-guided virtual staining.}

\begin{table}[!htbp]
  \centering
  \setlength\tabcolsep{2pt}
  \renewcommand{\arraystretch}{1.0}
  \caption{\revision{ABLATION STUDY OF PATHOLOGICAL GUIDANCE MECHANISM ON MIST DATASET. THE BEST SCORES ARE HIGHLIGHTED IN BOLD.}}
  \label{tab:ablation_foundation_1}

  \resizebox{\columnwidth}{!}{
    \begin{tabular}{c|cc|cc|cc}
      \Xhline{1.5pt} 
      \multirow{2}{*}{ Guidance} 
      & \multicolumn{2}{c|}{Quality} & \multicolumn{2}{c|}{Pathology} & \multicolumn{2}{c}{Perception} \\
      \cline{2-7}
      & PSNR$\uparrow$ & SSIM$\uparrow$ & IOD$_{\times10^7}$ & Pearson-R$\uparrow$ & FID$\downarrow$ & DISTS$\downarrow$ \\
      \hline
     PLIP & 14.2897 & \textbf{0.2111} & +0.6578 & 0.8476 & 41.4264 & 0.2511 \\
     MUSK & \textbf{14.7346} & 0.2089 & -2.5387 & 0.8408 & 40.6992 & 0.2499 \\
     CONCH & 14.3421 & 0.2032 & \textbf{-0.3223} & \textbf{0.8649} & \textbf{38.0654} & \textbf{0.2344} \\
      \Xhline{1.5pt} 
    \end{tabular}
  }
\end{table}

\subsection{Revisiting PSNR and SSIM as Evaluation Metrics for Virtual Staining}
\label{sec:discussion}
{
Specifically, in the virtual stain datasets (MIST and IHC4BC), the tissue samples for different stains originate from serial sections of the same specimen. Despite WSI-level registration, local tissue deformations and nonrigid morphological differences persist at the patch level. To further analyze the effect of image degradation, we simulate blurriness in the reference IHC images using Gaussian kernels of varying strengths. As observed by Zhu et al.~\cite{zhu2023breast}, introducing additional blur on top of already blurry IHC images can paradoxically increase the evaluation scores of virtual staining models (e.g., higher PSNR and SSIM), while the corresponding scores computed between the blurred labels and the original reference continue to decrease. This phenomenon highlights that conventional pixel-level metrics may overestimate performance when the outputs become perceptually smoother but biologically less faithful.
}

{
 To further illustrate this limitation, we evaluated PSNR and SSIM after applying Gaussian blur with different kernel sizes (Table~\ref{tab:combined_psnrssim_results}). Interestingly, increasing the blur level consistently improved both metrics, despite the apparent loss of fine structural details. This counterintuitive trend indicates that PSNR and SSIM are better suited as relative references rather than absolute indicators of perceptual or diagnostic quality in virtual staining tasks.
}

\begin{table}[h!]
\centering
\setlength\tabcolsep{1pt}
\caption{{COMPARISON OF PSNR AND SSIM VALUES WITH DIFFERENT GAUSSIAN BLUR KERNELS}}
\label{tab:combined_psnrssim_results}
\begin{tabular}{c|cccc|cccc}
\Xhline{1.5pt} 
\multirow{2}{*}{Method} & \multicolumn{4}{c|}{{PSNR}} & \multicolumn{4}{c}{{SSIM}} \\
\cline{2-9}
 & {No Blur} & {3×3} & {5×5} & {7×7} & {No Blur} & {3×3} & {5×5} & {7×7} \\
\hline
CycleGAN & 13.3060 & 13.4611 & 13.5518 & 13.6656 & 0.2000 & 0.2264 & 0.2401 & 0.2561 \\
Pix2Pix & 14.3445 & 14.4810 & 14.5597 & 14.6546 & 0.2143 & 0.2407 & 0.2558 & 0.2726 \\
CUT & 14.2183 & 14.3632 & 14.4553 & 14.5692 & 0.2047 & 0.2318 & 0.2466 & 0.2633 \\
ASP & 14.2746 & 14.4244 & 14.5189 & 14.6330 & 0.2144 & 0.2406 & 0.2553 & 0.2719 \\
Pyramid & 14.3320 & 14.4816 & 14.5731 & 14.6877 & 0.2137 & 0.2402 & 0.2554 & 0.2715 \\
TDKStain & 14.6360 & 14.7828 & 14.8702 & 14.9784 & 0.2181 & 0.2427 & 0.2563 & 0.2718 \\
PSPStain & 14.4773 & 14.6160 & 14.6973 & 14.8003 & 0.2206 & 0.2458 & 0.2605 & 0.2767 \\
DDBM & 14.2916 &14.3597 &14.4136 &14.4891 &0.2278 &0.2412 & 0.2511&0.2644\\
UNSB & 13.6369 &13.6961 &13.7460 &13.8196&0.2467 &0.2545 &0.2610&0.2702\\
ControlNet &9.1163 &9.1449 &9.1639 &9.1938 &0.1890 &0.1977 &0.2039 &0.2122\\
VIMs &13.7730 &14.0284&14.1404&14.2622&0.1848&0.2238&0.2436&0.2648\\
UMDST &13.4638 &13.5209&13.5664&13.6294&0.2455&0.2562&0.2638&0.2736\\

PGVMS & 14.3421 & 14.5022 & 14.6046 & 14.7304 & 0.2032 & 0.2298 & 0.2453 & 0.2624 \\
\hline
Label & - & 36.4453 & 33.0669 & 30.3883 & - & 0.9627 & 0.9189 & 0.8530 \\
\Xhline{1.5pt} 
\end{tabular}
\end{table}

\subsection{Training Efficiency and Scalability}
{
PGVMS serves as a unified model for multiplex staining, maintaining a compact architecture with a model size of 26.35 MB and requiring 16,540 MB of GPU memory during training on 512×512 patches cropped from 1024×1024 whole-slide regions. The training process completes within 125 hours across four stain datasets for 80 epochs.}
\begin{table}[h!]
\centering
\caption{{Computational efficiency of PGVMS during training and inference.}}
\label{tab:pgvms_efficiency}
\resizebox{\columnwidth}{!}{%
\begin{tabular}{lccc}
\toprule
\textbf{Stage} & \textbf{Input Size} & \textbf{GPU (MB)} & \textbf{Time} \\
\midrule
Training & $512\times512$ & 16,540 & 125 h / 80 epochs  \\
Inference & $1024\times1024$ & 10,284 & 0.0208 s / tile  \\
\bottomrule
\end{tabular}%
}
\end{table}

{
During inference, PGVMS achieves high throughput and scalability on whole-slide images (WSIs). 
For 1024×1024 tiles, the model processes each tile in 0.0208 seconds using 10,284 MB GPU memory. 
}

{
These results (Table \ref{tab:pgvms_efficiency}) demonstrate that PGVMS is computationally efficient, has a low memory footprint, 
and scales well for large WSIs, making it suitable for clinical virtual staining applications.
}



    
    
    


\subsection{Prompt Robustness Evaluation of PGVMS}

{
We evaluated PGVMS under diverse prompt formulations to assess its robustness to linguistic variability. For each biomarker (HER2, ER, PR, Ki67), six prompts were designed and categorized by length and semantic richness—ranging from concise functional descriptions (\textit{short}, e.g., \textit{\revision{``H\&E to HER2/ER/PR/Ki67 stain conversion"}}), to moderately descriptive expressions (\textit{medium}, e.g., \textit{\revision{``Convert H\&E histology slide to HER2/ER/PR/Ki67 staining"}}), and semantically enriched, context-aware formulations (\textit{long}, e.g., \textit{\revision{``Convert this H\&E WSI patch to a HER2/ER/PR/Ki67 stain, ensuring the highest possible fidelity and biological plausibility"}}). In addition, we included the \textit{original} prompt (\textit{\revision{``From an H\&E stained image transfer to a HER2/ER/PR/Ki67 stained image"}}) as a reference, consistent with the default training configuration.
}


\revision{Table~\ref{tab:prompt_robustness} reports the quantitative comparison under different prompt formulations. 
Following prior prompt-guided virtual staining studies such as VIMs\cite{dubey2024vims}, we analyze prompt robustness by categorizing prompts according to their length and semantic informativeness, reflecting increasing levels of linguistic and biological conditioning. 
Overall, PGVMS exhibits only minor and acceptable performance variations across  short ($\sim$5 words), medium ($\sim$10 words), and long ($\sim$18 words) prompts, indicating that the model is not overly sensitive to prompt formulation.
}


\begin{table}[ht]
  \centering
  \setlength\tabcolsep{2pt}
  \renewcommand{\arraystretch}{1.0}
  \caption{{ROBUSTNESS EVALUATION OF PGVMS UNDER DIFFERENT PROMPT EXPRESSIONS. THE BEST SCORES ARE HIGHLIGHTED IN BOLD.}}
  \label{tab:prompt_robustness}

  \resizebox{\columnwidth}{!}{
    \begin{tabular}{cc|cc|cc|cc}
      \Xhline{1.5pt} 
      \multirow{2}{*}{Length} & \multirow{2}{*}{Prompt} 
      & \multicolumn{2}{c|}{Quality} & \multicolumn{2}{c|}{Pathology} & \multicolumn{2}{c}{Perception} \\
      \cline{3-8}
      & & PSNR$\uparrow$ & SSIM$\uparrow$ & IOD$_{\times10^7}$ & Pearson-R$\uparrow$ & FID$\downarrow$ & DISTS$\downarrow$ \\
      \hline
      Short & 1 & 14.4628 & 0.2211 & -0.9492 & 0.8284 & 44.6465 & 0.2420 \\
      Short & 2 & 14.2855 & 0.2092 & -0.5466 & 0.8433 & 38.4802 & 0.2385 \\
      \cdashline{1-8}[0.8pt/2pt]
      Medium & 3 & 14.2589 & \textbf{0.2333} & -1.7759 & 0.7235 & 53.0694 & 0.2541 \\
      Medium & 4 & 14.4086 & 0.2160 & -1.4156 & 0.8326 & 40.1634 & 0.2410 \\
      \cdashline{1-8}[0.8pt/2pt]
      Long & 5 & 14.4193 & 0.2245 & -1.2309 & 0.7994 & 46.1909 & 0.2441 \\
      Long & 6 & \textbf{14.4716} & 0.2300 & -1.8585 & 0.7977 & 46.4370 & 0.2454 \\
      \cdashline{1-8}[0.8pt/2pt]
      Origin & - & 14.3421 & 0.2032 & \textbf{-0.3223} & \textbf{0.8649} & \textbf{38.0654} & \textbf{0.2344} \\
      \Xhline{1.5pt} 
    \end{tabular}
  }
\end{table}

\revision{Consistent with observations in VIMs~\cite{dubey2024vims}, increasing prompt length or semantic richness does not necessarily translate into improved virtual staining quality, and may even introduce slight degradations in certain metrics. 
 This suggests that overly complex descriptions can introduce redundant or noisy semantic cues. 
By leveraging the CONCH-guided embedding, PGVMS is able to extract domain-relevant semantic information from concise and moderately informative prompts without relying on overly verbose descriptions, thereby preserving biological plausibility across diverse prompt formulations.}


\subsection{Generalization to Unseen Clinical Breast Cancer Patients}
{
To demonstrate the generalization ability of our method to unseen patients, 
we evaluated PGVMS on a private clinical breast cancer dataset. 
A small subset of this dataset (9 ROIs) contains all four IHC stains (HER2, ER, PR, Ki-67), 
with each image cropped from whole-slide images (WSIs) into patches of size $4096 \times 3072$.
}

\begin{figure}[!htbp]
    \centering
    \includegraphics[width=\columnwidth,height=1\textheight,keepaspectratio]{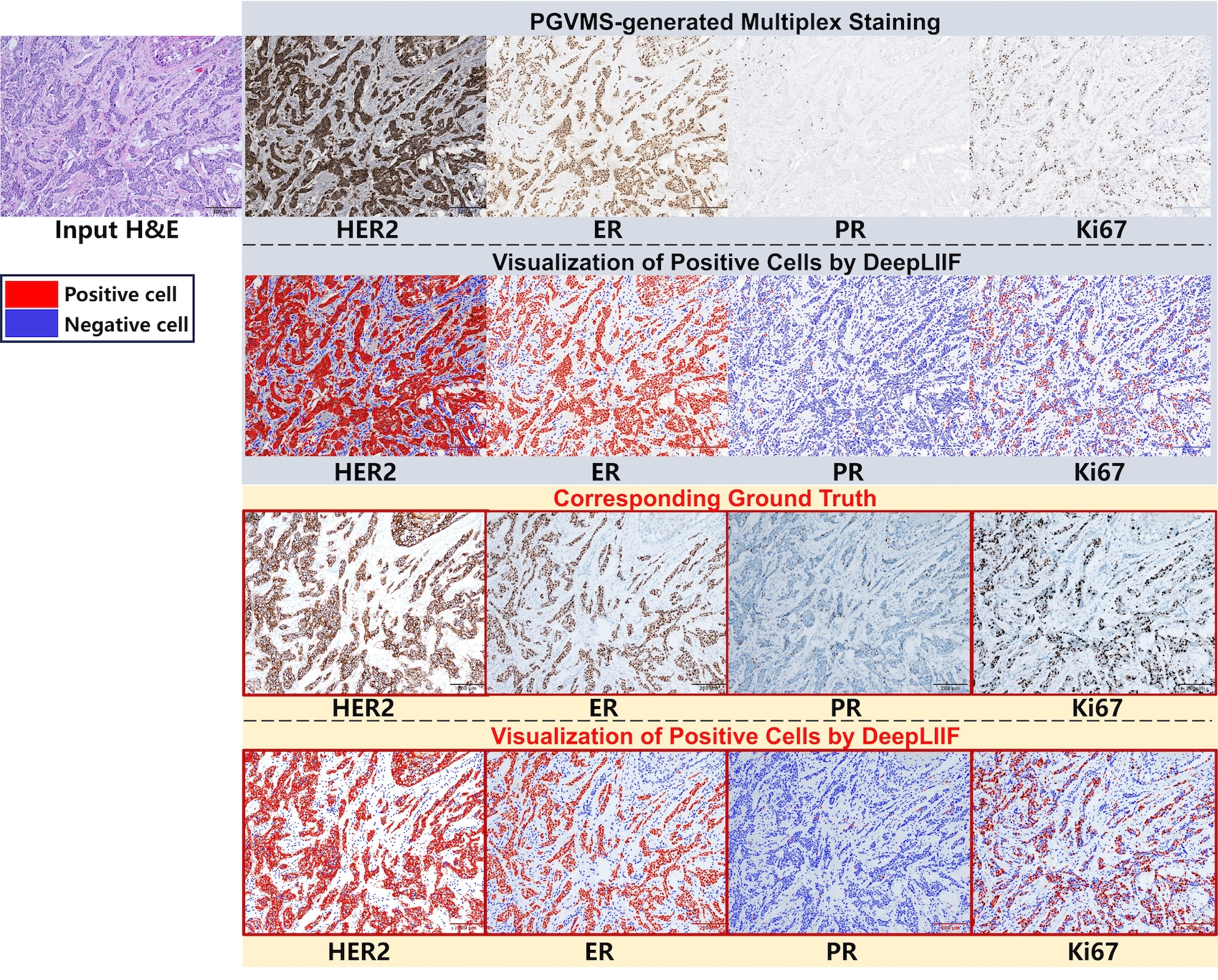}
    \caption{\revision{Virtual IHC-stained image results on the clinical dataset, showing four stains (HER2, ER, PR, Ki-67) and positive cell visualization obtained using DeepLIIF. The first column shows the input H\&E images. Images with a gray background represent PGVMS-generated multiplex stains along with their corresponding positive cell visualization, while images with an yellow background show the ground truth multiplex stains and the associated positive cell visualization.}}

    \label{qualitative_qin_bci_2}
\end{figure}

\begin{figure}[!htbp]
    \centering
    \includegraphics[width=\columnwidth,height=1\textheight,keepaspectratio]{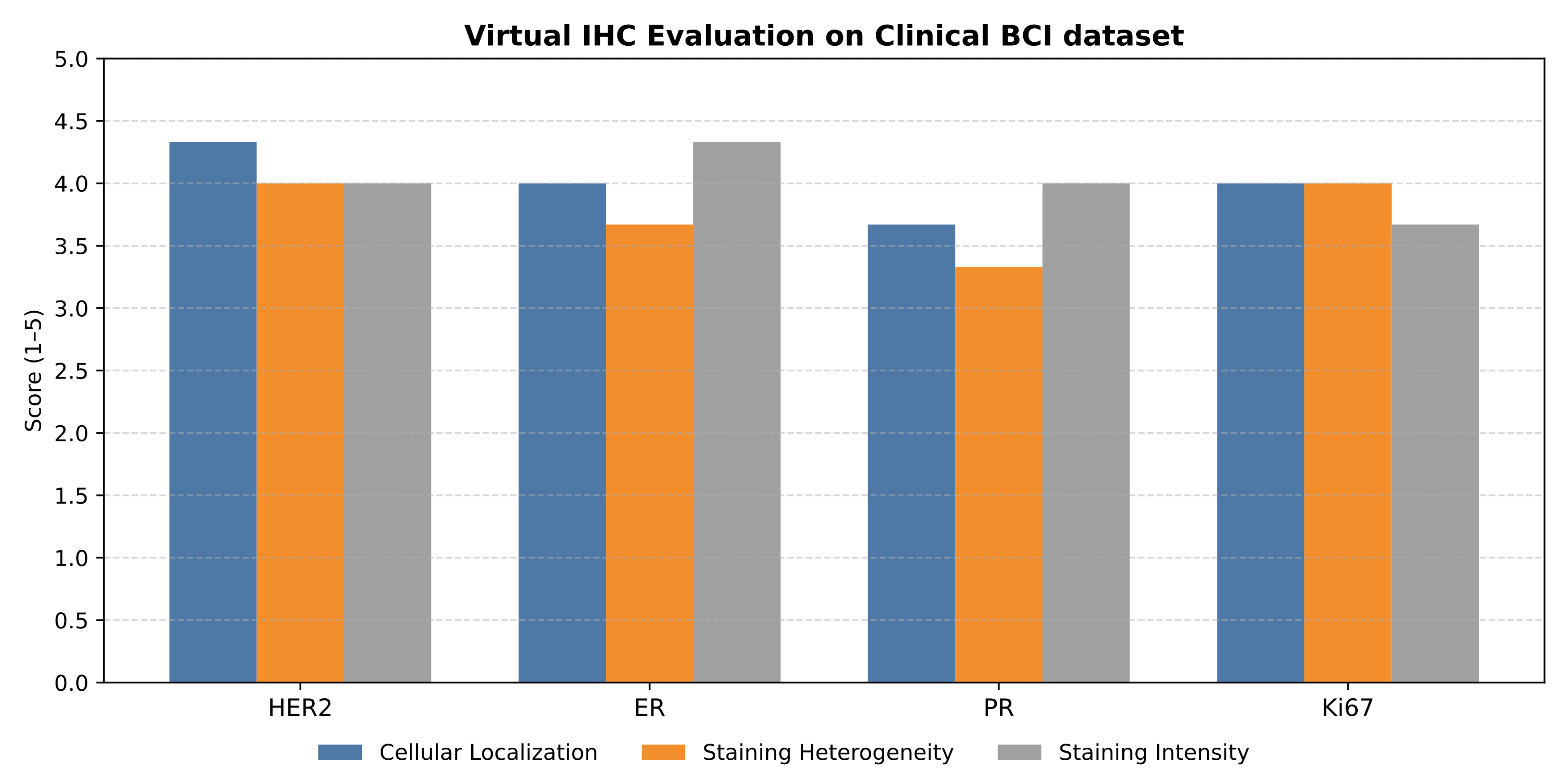}
    \caption{{Virtual IHC-stained image results on the clinical dataset.  PGVMS preserves accurate cellular localization and realistic staining intensity across different biomarker.}}
    \label{qualitative_qin_bci_3}
\end{figure}

{
As shown in Figure~\ref{qualitative_qin_bci_2}, our method robustly generates all four stains for previously unseen patients, 
demonstrating strong generalization at the patient level. An experienced pathologist evaluated the virtual IHC-stained images on a five-point scale (1–5) across three criteria: cellular localization, staining heterogeneity, and staining intensity, as shown in Figure~\ref{qualitative_qin_bci_3}.
These results indicate that PGVMS can effectively handle variations in tissue morphology and staining patterns, supporting its potential applicability in diverse clinical scenarios.
}

\subsection{Cross-Organ and Cross-Cancer Generalization of PGVMS}

{
To evaluate the generalization ability of our PGVMS framework across different organs, cancer types, and IHC biomarkers, 
we conducted experiments on multiple independent clinical datasets, covering colon, liver, nasopharyngeal, and head-and-neck cancers. 
}
{
For each organ, we collect representative ROIs containing key biomarkers (e.g., Ki-67 for colorectal, HepPar1 for liver,  EBER and CK for nasopharyngeal carcinoma, PD-L1 for head and neck).
}




\subsubsection{Clinical Colon Cancer Dataset} 
{
We collected 69 ROIs from WSIs, each with paired upper and lower sections stained with H\&E and Ki-67 IHC.  
ROIs were cropped to $4096 \times 3072$ patches, and transfer learning for 10 epochs was applied.
}

\begin{figure}[!htbp]
\centerline{\includegraphics[width=\columnwidth,height=1\textheight,keepaspectratio]{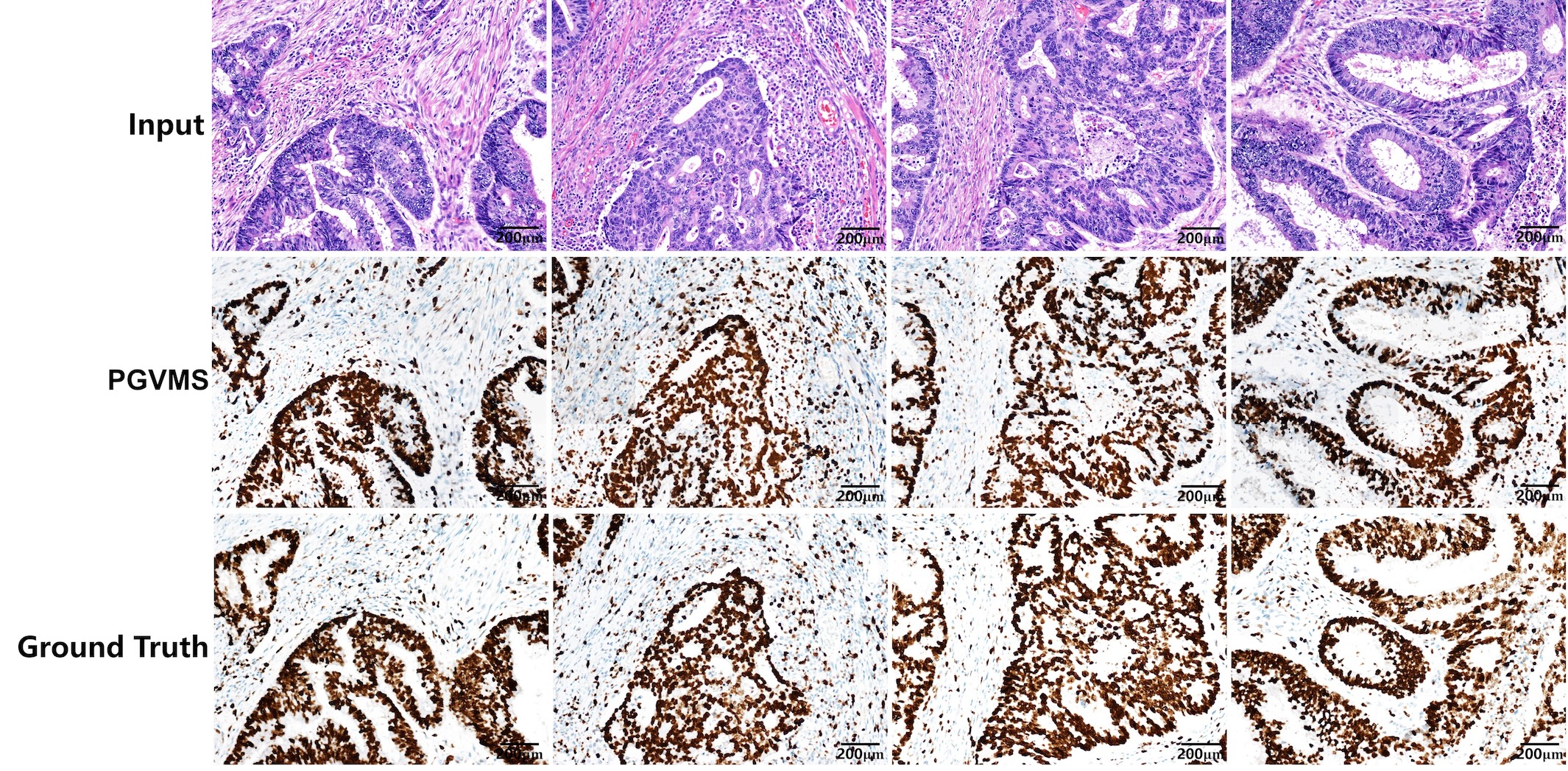}}
\caption{{Virtual Ki-67-stained image results on the clinical colon cancer dataset.}}
\label{qualitative_colon_1}
\end{figure}

{
PGVMS synthesizes Ki-67-stained images (Fig. \ref{qualitative_colon_1}) with clear nuclear localization and structural fidelity comparable to the reference IHC slides.}

\subsubsection{Clinical Hepatocellular Carcinoma (HCC) Dataset} 
{
38 ROIs with paired H\&E and HepPar1 IHC sections were cropped to $4096 \times 3072$, and transfer learning was applied for 10 epochs.
}

\begin{figure}[!htbp]
\centerline{\includegraphics[width=\columnwidth,height=1\textheight,keepaspectratio]{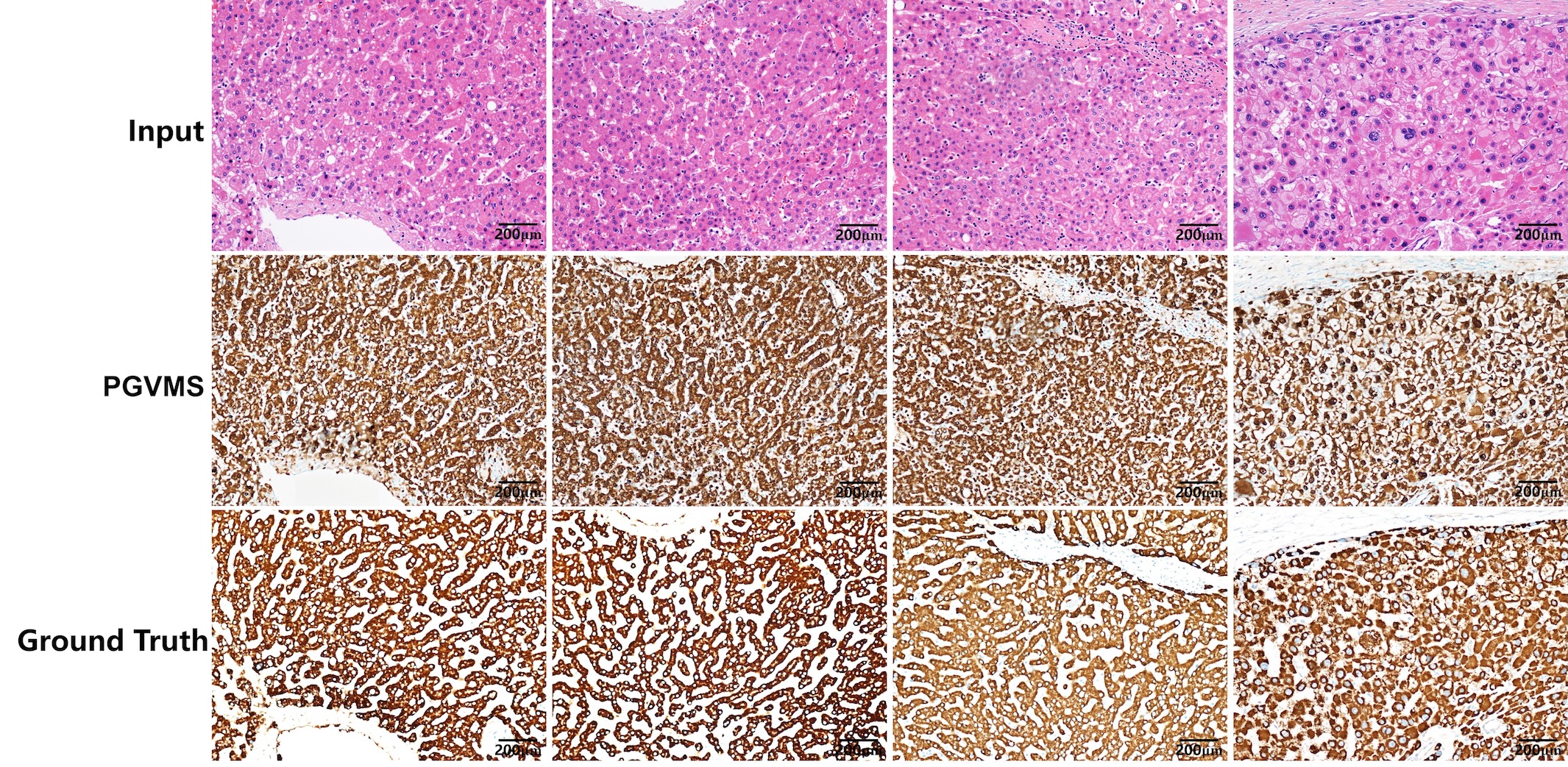}}
\caption{{Virtual HepPar1 IHC-stained image results on the clinical HCC dataset.}}
\label{qualitative_liver_1}
\end{figure}

{
PGVMS (Fig. \ref{qualitative_liver_1}) preserves hepatocyte morphology and cytoplasmic staining patterns consistent with ground-truth slides.
}

\subsubsection{Clinical Nasopharyngeal Carcinoma (NPC) Dataset} 
{
57 ROIs were collected: 46 cases with CK and EBER stains, 11 cases with CK only. ROIs were cropped to $4096 \times 3072$, and the pretrained model was fine-tuned for 10 epochs.
}

\begin{figure}[!htbp]
\centerline{\includegraphics[width=\columnwidth,height=1\textheight,keepaspectratio]{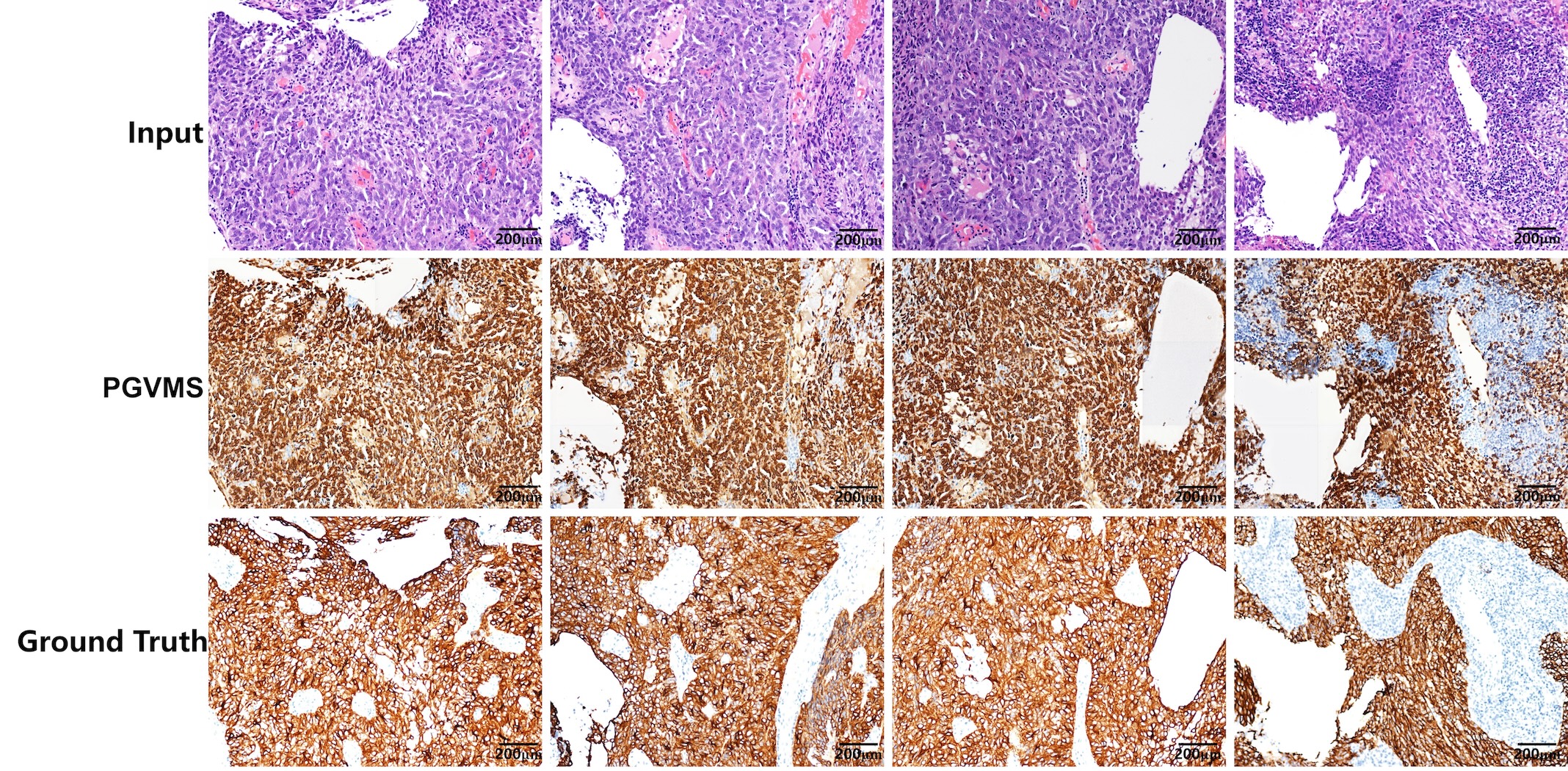}}
\caption{{Virtual CK-stained image results on the clinical NPC dataset.}}
\label{qualitative_CK_1}
\end{figure}

\begin{figure}[!htbp]
\centerline{\includegraphics[width=\columnwidth,height=1\textheight,keepaspectratio]{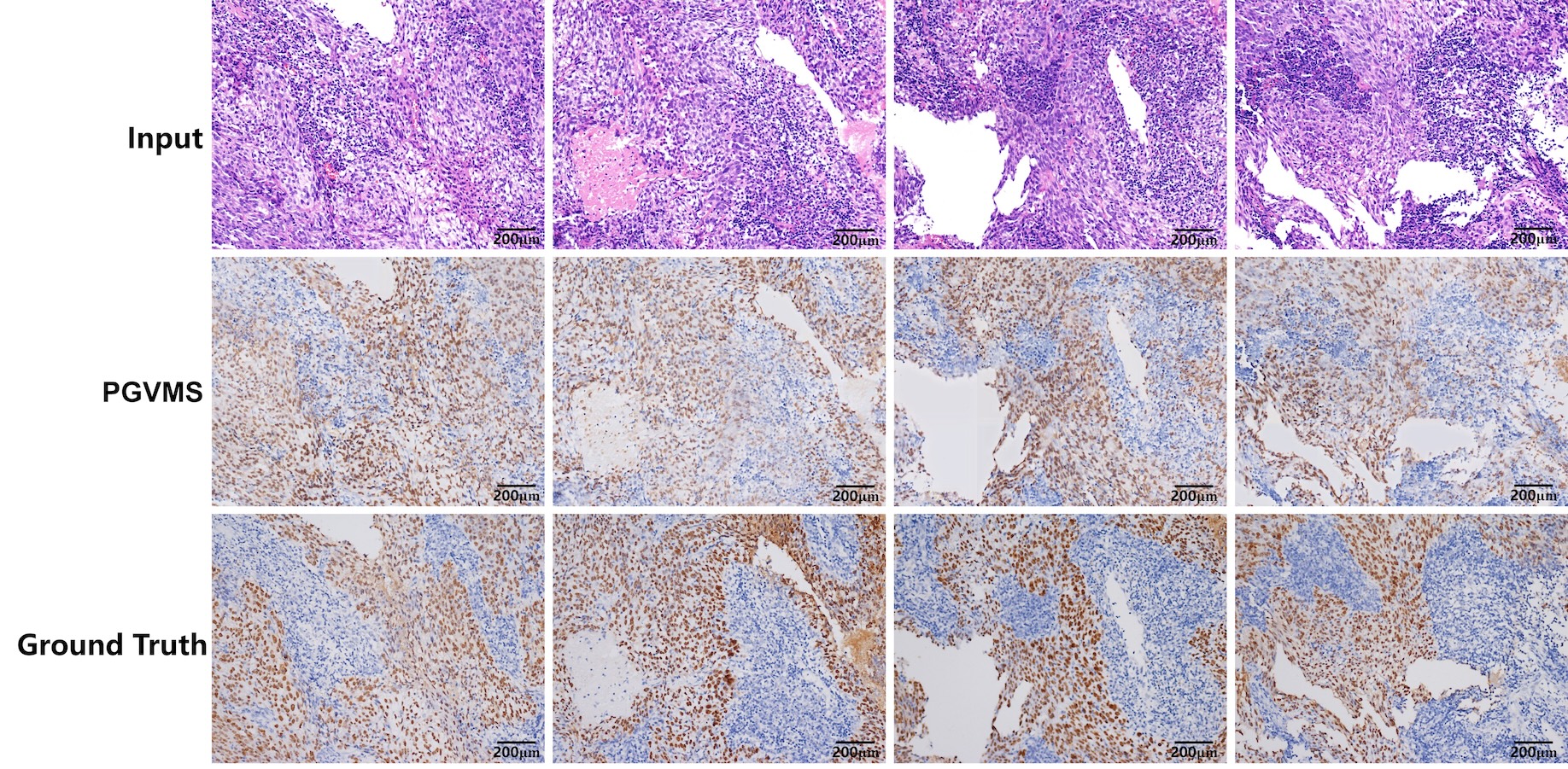}}
\caption{{Virtual EBER-stained image results on the clinical NPC dataset.}}
\label{qualitative_EBER_1}
\end{figure}

{
PGVMS synthesizes CK and EBER stains (Fig. \ref{qualitative_CK_1} and Fig. \ref{qualitative_EBER_1}) with realistic contrast and structural fidelity comparable to reference slides.
}

\subsubsection{Clinical Head and Neck Squamous Cell Carcinoma (HNSCC) Datasets} 
{
17 ROIs with paired H\&E and PD-L1 IHC sections were cropped to $4096 \times 3072$ patches, and the model was fine-tuned for 10 epochs.
}

\begin{figure}[!htbp]
\centerline{\includegraphics[width=\columnwidth,height=1\textheight,keepaspectratio]{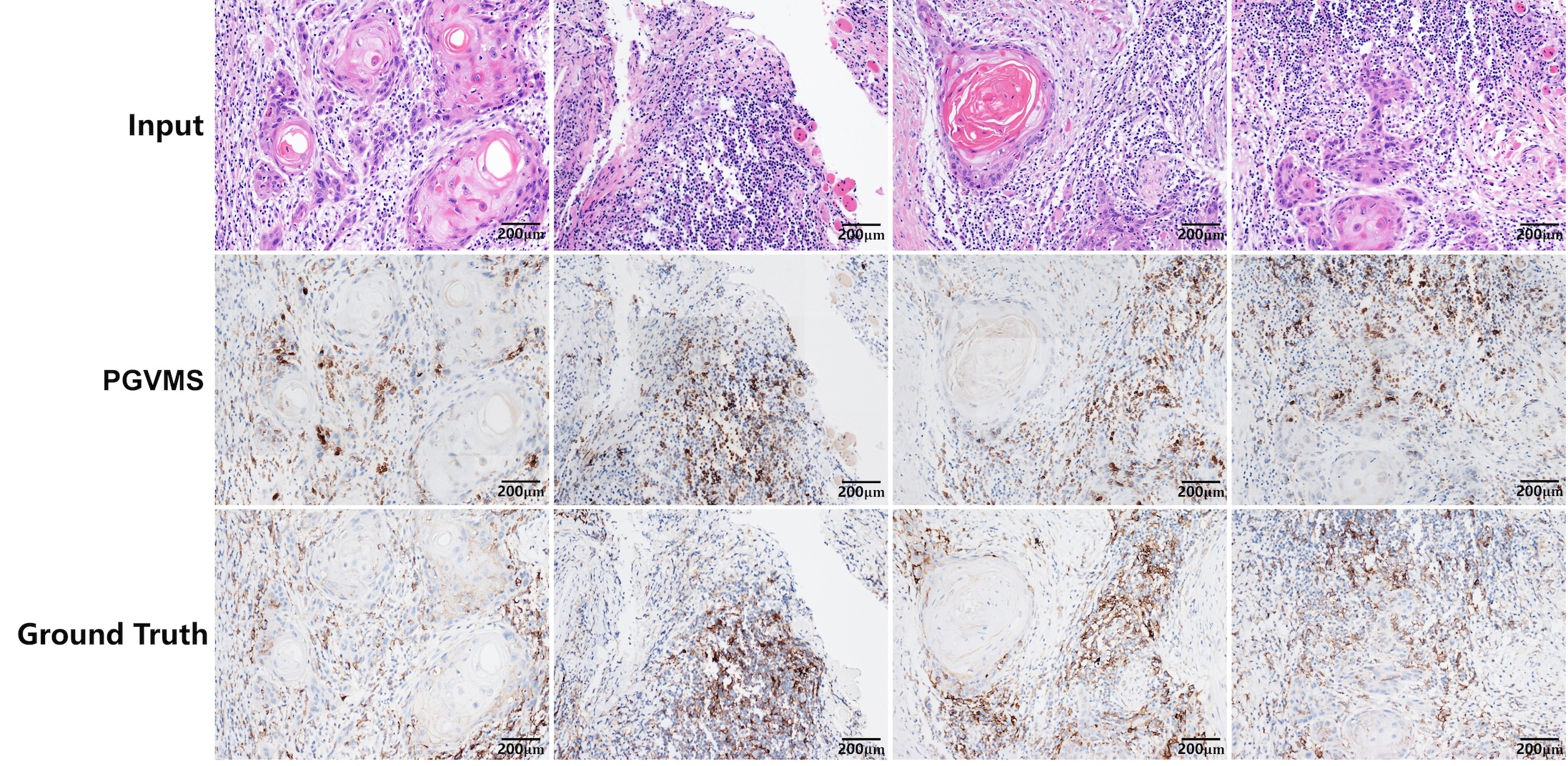}}
\caption{{Virtual PD-L1 IHC-stained image results on the clinical HNSCC dataset.}}
\label{qualitative_hnscc_1}
\end{figure}

{
PGVMS generates virtual PD-L1-stained images (Fig. \ref{qualitative_hnscc_1}) exhibiting realistic membrane staining and expression intensity consistent with reference slides.
}

\subsubsection{Pathologist Evaluation}
\begin{figure}[!htbp] \centerline{\includegraphics[width=\columnwidth]{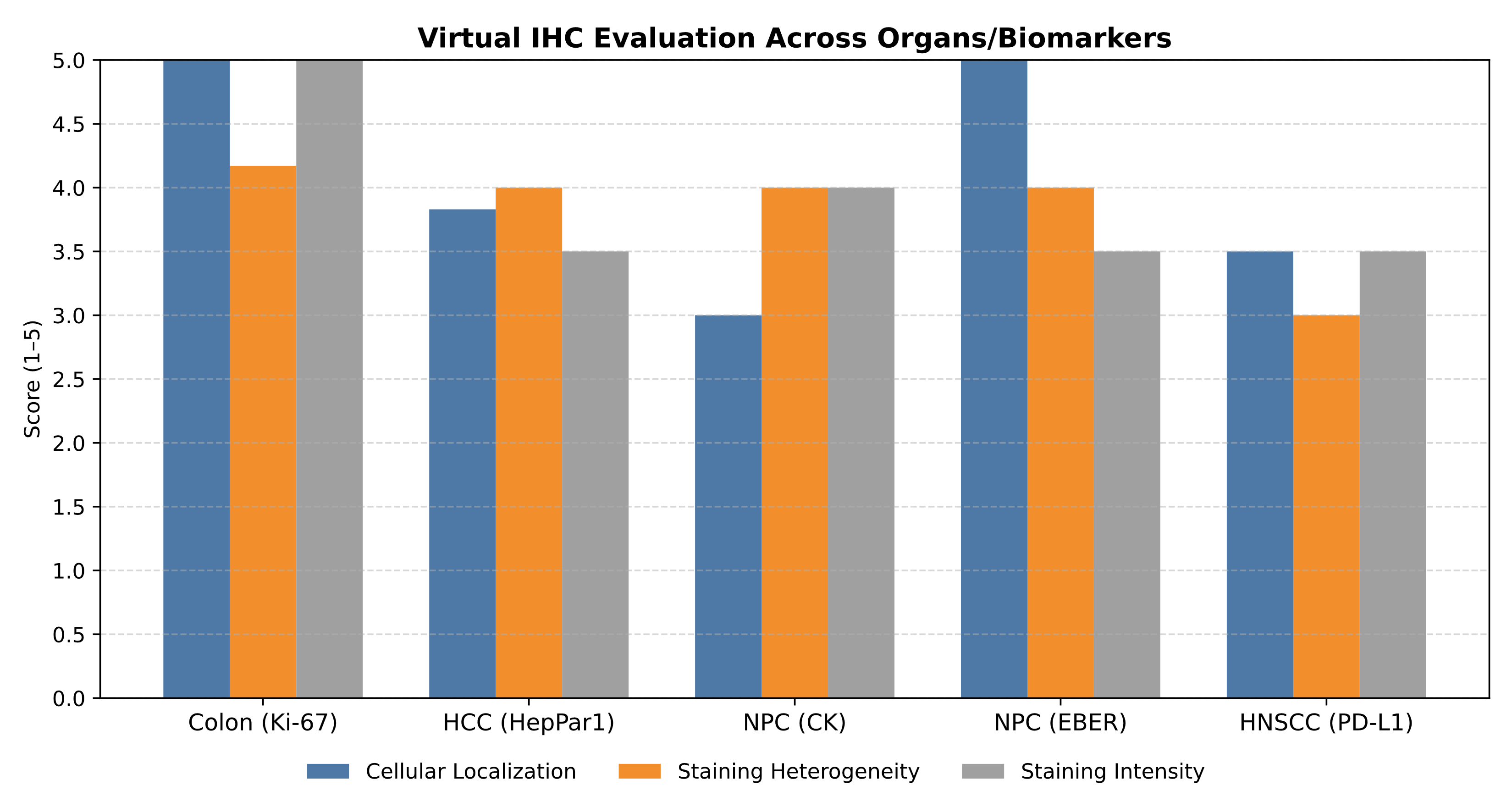}} 
\caption{{Representative cross-organ virtual IHC staining results on liver, head-and-neck, colorectal, and nasopharyngeal samples. 
PGVMS preserves accurate cellular localization and realistic staining intensity across different biomarkers.}}
\label{qualitative_organ} 
\end{figure}

{Experienced pathologist was asked to evaluate the virtual IHC-stained images on a five-point scale (1–5) across three aspects: 
cellular localization, {staining heterogeneity}, and {staining intensity} in Fig.~\ref{qualitative_organ}. 
}

{
Specifically, Ki-67 and EBER achieve the highest overall ratings, demonstrating accurate nuclear localization and strong contrast between positive and negative regions. 
HepPar1 also performs well, showing clear cytoplasmic staining with minimal artefacts. 
PD-L1 and CK exhibit stable results across all three aspects, confirming the robustness of PGVMS on diverse tissue structures and staining patterns. 
Across all datasets, PGVMS demonstrates strong cross-organ and cross-cancer generalization, accurately synthesizing multiple IHC stains from H\&E images, maintaining morphological fidelity, and preserving biologically meaningful staining patterns suitable for diverse clinical scenarios.
}

\section{Conclusion}
In this work, we present PGVMS, a comprehensive framework for pathological prompt-guided IHC virtual multiplex staining using only uniplex training data. Our three key innovations: (1) the pathological semantics-style guided  (PSSG) generator with CONCH integration, (2) protein-aware learning strategy (PALS) with focal optical density mapping and (3) prototype-consistent learning strategy (PCLS) collectively address the critical challenges of semantic preservation, molecular-level accuracy, and spatial alignment in stain transformation. \revision{Extensive validation on MIST, IHC4BC, and cross-organ clinical datasets demonstrates state-of-the-art performance, with our method outperforming existing generative approaches while requiring no additional annotations.} This work bridges a critical gap in computational pathology by evolving virtual staining from isolated single-task solutions to an integrated system supporting multiplex diagnosis via prompt-based control. 

{
Our method, through the PCLS module, performs particularly well on clustered IHC patterns. However, biomarkers such as Ki-67, which display more dispersed staining, may be less optimally captured, whereas regions with contiguous positive cells are handled more effectively. Additionally, the framework does not explicitly incorporate biomarker-specific spatial priors. For instance, HER2 is membrane-bound, ER/PR are nuclear, and Ki-67 labels proliferating nuclei. Incorporating such localization constraints could further enhance biological realism and produce virtual staining results that are more clinically plausible.
}

{
In future work, we plan to investigate strategies to better handle dispersed IHC staining patterns and to explore the scalability of PGVMS for multiplex IHC applications. By explicitly leveraging heterogeneous spatial information across different IHC stains, we aim to improve the biological realism, robustness, and clinical utility of virtual multiplex staining.
}

\bibliographystyle{IEEEtran}
\bibliography{ref.bib}


\end{document}